\documentclass[journal]{IEEEtran}
\usepackage{cite}

\ifCLASSINFOpdf
\else
\fi
\usepackage{amsmath}
\usepackage{amssymb}
\usepackage{amsfonts}
\usepackage{bm}
\usepackage{algorithmic}
\usepackage{array}
\usepackage{url}
\usepackage{booktabs}
\usepackage{multirow}
\usepackage{graphicx}
\usepackage{colortbl}
\usepackage{xcolor}

\definecolor{mediumgray}{gray}{0.88}
\definecolor{lightgray}{gray}{0.95}

\newcommand{\best}[1]{\textbf{#1}}
\newcommand{\second}[1]{\underline{#1}}

\DeclareMathOperator*{\E}{\mathbb{E}}

\begin{document}
%
\title{WindFM: An Open-Source Foundation Model for Zero-Shot Wind Power Forecasting}
%
%
%
\author{Hang Fan\textsuperscript{\dag},~\IEEEmembership{Member,~IEEE,}
        Yu Shi\textsuperscript{\dag}, Zongliang Fu, Shuo Chen, Wei Wei,~\IEEEmembership{Member,~IEEE},
        Wei Xu, and Jian Li
\thanks{Hang Fan is with the School of Economics and Management, North China Electric Power University, Beijing 102206, China (e-mail: fanhang123456@163.com). Yu Shi, Zongliang Fu, Shuo Chen, Jian Li, Wei Xu, Wei Wei are with Tsinghua University, Beijing 100084, China. Yu Shi, Shuo Chen, Wei Xu and Jian Li are also with the Institute for Interdisciplinary Information Sciences. Zongliang Fu is also with the Department of Automation. Wei Wei is also with the Department of Electrical Engineering.}
\thanks{\textsuperscript{\dag}Hang Fan and Yu Shi are equal contribution to this work.}
}

\maketitle

\begin{abstract}
High-quality wind power forecasting is crucial for the operation of modern power grids. However, prevailing data-driven paradigms either train a site-specific model
 which cannot generalize to other locations or rely on fine-tuning of general-purpose time series foundation models which are difficult to incorporate domain-specific data in the energy sector. This paper introduces WindFM, a lightweight and generative Foundation Model designed specifically for probabilistic wind power forecasting. WindFM employs a discretize-and-generate framework. A specialized time-series tokenizer first converts continuous multivariate observations into discrete, hierarchical tokens. Subsequently, a decoder-only Transformer learns a universal representation of wind generation dynamics by autoregressively pre-training on these token sequences. Using the comprehensive WIND Toolkit dataset comprising approximately 150 billion time steps from more than 126,000 sites, WindFM develops a foundational understanding of the complex interplay between atmospheric conditions and power output. Extensive experiments demonstrate that our compact 8.1M parameter model achieves state-of-the-art zero-shot performance on both deterministic and probabilistic tasks, outperforming specialized models and larger foundation models without any fine-tuning. In particular, WindFM exhibits strong adaptiveness under out-of-distribution data from a different continent, demonstrating the robustness and transferability of its learned representations. Our pre-trained model is publicly available at \url{https://github.com/shiyu-coder/WindFM}.
\end{abstract}


%
\IEEEpeerreviewmaketitle

\section{Introduction}
\IEEEPARstart{T}{o} address the pressing challenge of climate change, the installed capacity of renewable energy sources has witnessed significant growth globally \cite{gwec2024}. Among these, wind power has emerged as a cornerstone of the clean energy transition \cite{wwindea2024}. However, the large-scale integration of wind power introduces substantial volatility and uncertainty into the electric grid due to its inherent dependence on fluctuating meteorological conditions. Consequently, high-quality wind power forecasting across multiple time scales is indispensable for the energy sector, underpinning the secure and economic operation of modern power systems and the effective participation of wind power producers in electricity markets \cite{wang2024high}. Improving the precision of forecasts is critical for maximizing wind energy utilization, ensuring grid stability, and enhancing the profitability of wind farms.

\begin{figure}[t]
    \centering
    \includegraphics[width=\columnwidth]{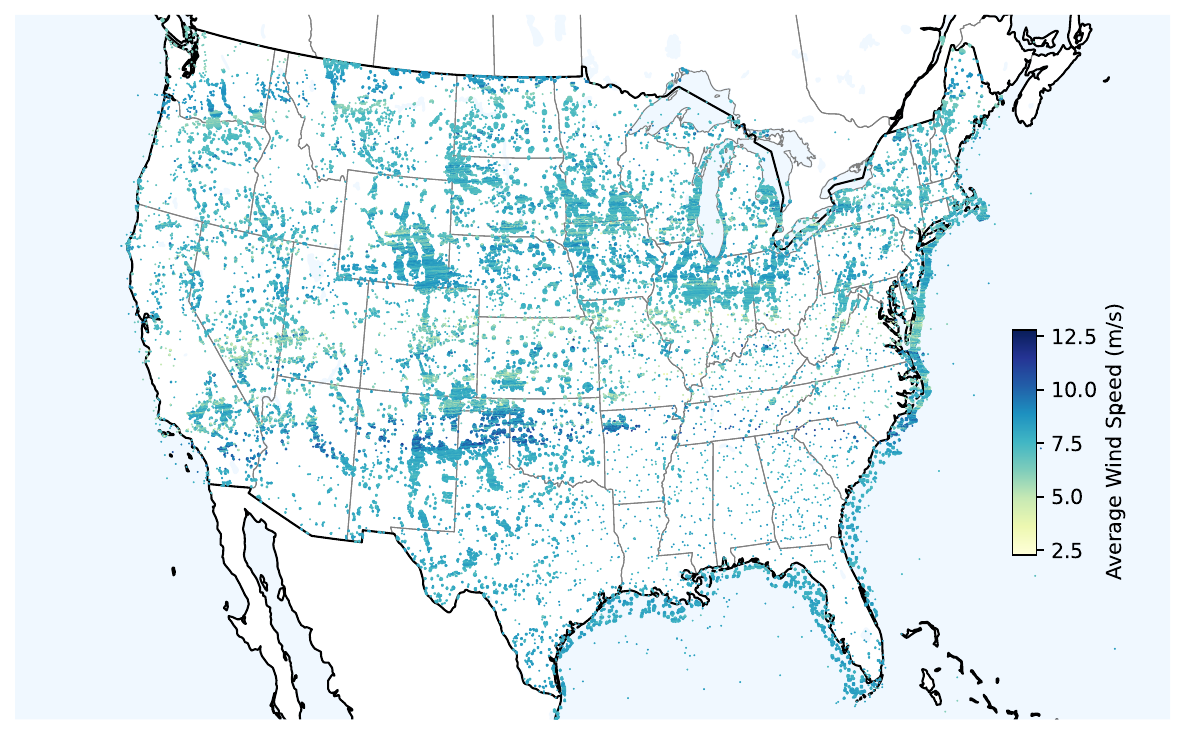}
    \caption{Geographical distribution of wind turbine sites in the WIND Toolkit dataset across the continental United States. The color of each point represents the average annual wind speed at the site, indicating regions with high wind energy potential.}
    \label{fig:station_map}
\end{figure}

\begin{figure*}[t!]
    \centering
    \includegraphics[width=1.0\textwidth]{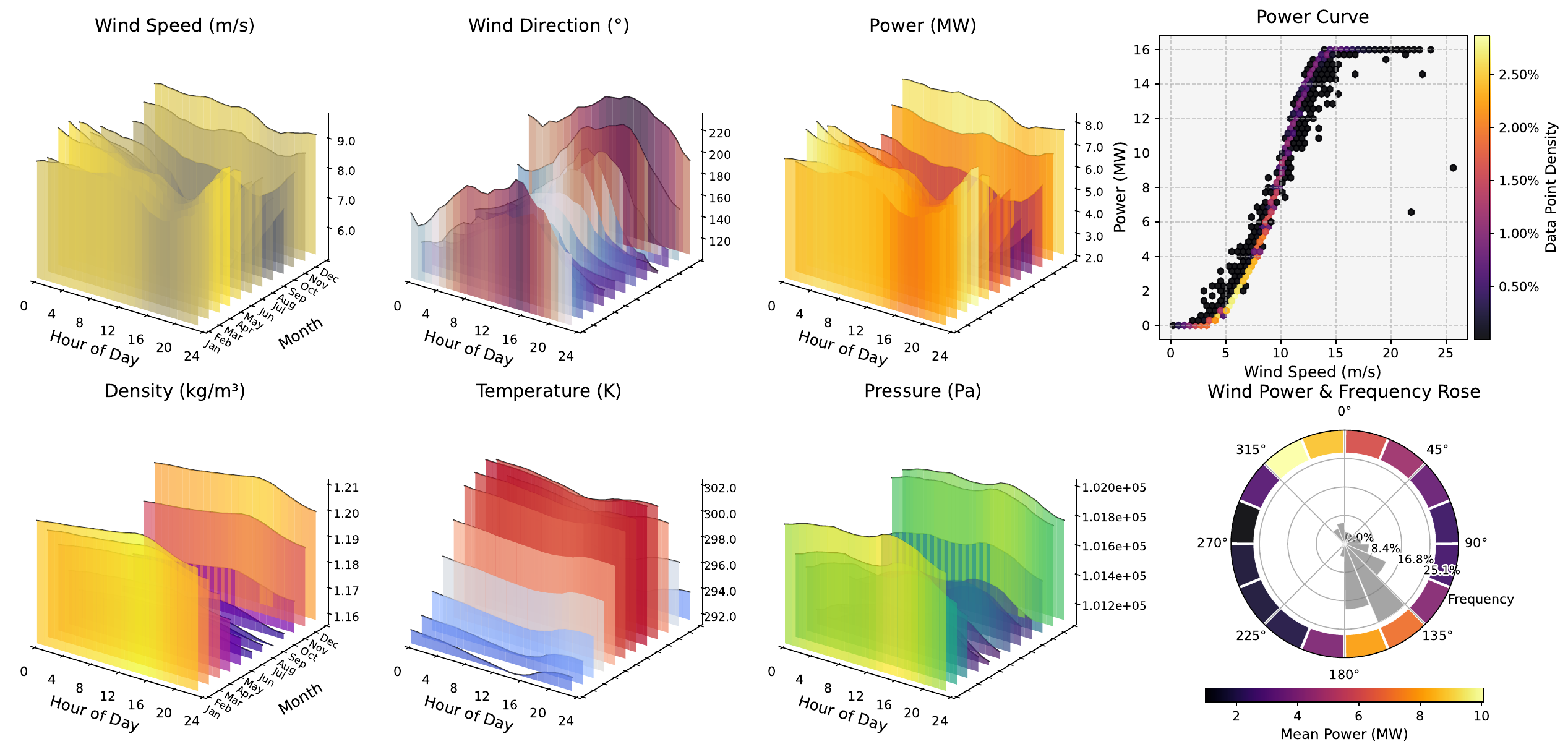}
    \caption{Visualization of complex wind dynamics from a representative site. (Left) Six 3D plots reveal strong diurnal and seasonal cycles in meteorological conditions and power generation. (Top Right) The empirical power curve shows the non-linear relationship between wind speed and power, with color representing observation density. (Bottom Right) A composite wind rose visualizes the frequency of wind direction (bar length) and the associated mean power (segment color). These characteristics highlight the intricate, multi-scale dependencies that necessitate a powerful data-driven forecasting model.}
    \label{fig:time_variation_pattern}
\end{figure*}

Existing wind power prediction methods are broadly categorized into 
physics-based
and data-driven approaches \cite{lu2021review}. Physical models, which rely on numerical weather prediction (NWP) and turbine power curves, offer high interpretability but are often limited in accuracy by coarse spatiotemporal resolutions and sensitivity to initial and boundary conditions \cite{zhang2024temporal,li2025novel}. In contrast, with the proliferation of sensor data and advances in artificial intelligence, data-driven methods have become the prevailing paradigm, demonstrating superior accuracy by learning complex patterns directly from historical data \cite{yang2024survey}. This category includes classical statistical models as well as contemporary deep learning architectures such as Recurrent Neural Networks (RNNs) and Transformers \cite{shahid2021novel, huang2023deep, wang2024high, zhang2021multi,cheng2021augmented,wen2022continuous}. More recently, techniques inspired by large language models (LLMs) have also been explored for energy forecasting applications \cite{zhou2025empower,majumder2024exploring,jiang2024eplus,lai2024bert4st}.

Despite their success, existing data-driven models face fundamental limitations. First, most are trained in a site-specific manner, requiring a separate model to be developed and maintained for each wind farm \cite{zhang2024temporal}. This approach is not only computationally expensive and labor-intensive but also fails to leverage the rich, shared dynamics and weather patterns that exist across diverse geographical locations. Second, while recent general-purpose time-series foundation models (e.g., Chronos \cite{ansari2024chronos}, TimeMOE \cite{shi2024time}) offer a potential alternative, their large parameter counts can hinder practical deployment in operational environments. Furthermore, as demonstrated in our comparative analysis, they often struggle to outperform well-tuned, specialized models in zero-shot forecasting scenarios without requiring extensive, site-specific fine-tuning.

To overcome these challenges, we propose a domain-specific foundation model, an approach whose efficacy is supported by recent successes in various specialized domains, such as electricity pricing \cite{yu2025pricefm} and finance \cite{shi2025kronos}. Building on this principle, we introduce the Wind Power Foundation Model (WindFM), a generative framework tailored for probabilistic wind power forecasting. By pre-training a single, unified model on vast, multi-site meteorological and power data, WindFM learns a universal representation of the complex interplay between atmospheric conditions and turbine generation. Our approach employs a discretize-and-generate framework, where a specialized time-series tokenizer converts continuous observations into discrete tokens, and a decoder-only Transformer learns the underlying dynamics of the resulting token sequences. This paradigm offers a scalable and efficient solution that excels at zero-shot forecasting, providing high-quality predictions for unseen locations without any model retraining or fine-tuning. The main contributions of this work are twofold:
\begin{itemize}
\item We propose the architecture of WindFM, a lightweight foundation model tailored for wind power forecasting. With only 8.1M parameters, this compact architecture learns generalizable representations while facilitating easy deployment in practical energy system applications.
\item We perform, to the best of our knowledge, the first large-scale pre-training of a foundation model exclusively on wind energy data. Using the comprehensive WIND Toolkit \cite{draxl2015wind}, our model learns from a massive dataset comprising approximately 150 billion multivariate time steps from more than 126,000 sites, capturing a diverse range of meteorological conditions, operational dynamics, and temporal resolutions.
\end{itemize}

The effectiveness of WindFM is validated through extensive experiments on unseen domestic sites and a challenging out-of-distribution dataset from wind farms in Inner Mongolia, China. We demonstrate that WindFM achieves state-of-the-art zero-shot performance on both deterministic and probabilistic forecasting tasks, outperforming specialized models and larger foundation models without any fine-tuning.

\section{Preliminary}

\begin{figure*}[t!]
    \centering
    \includegraphics[width=0.9\textwidth]{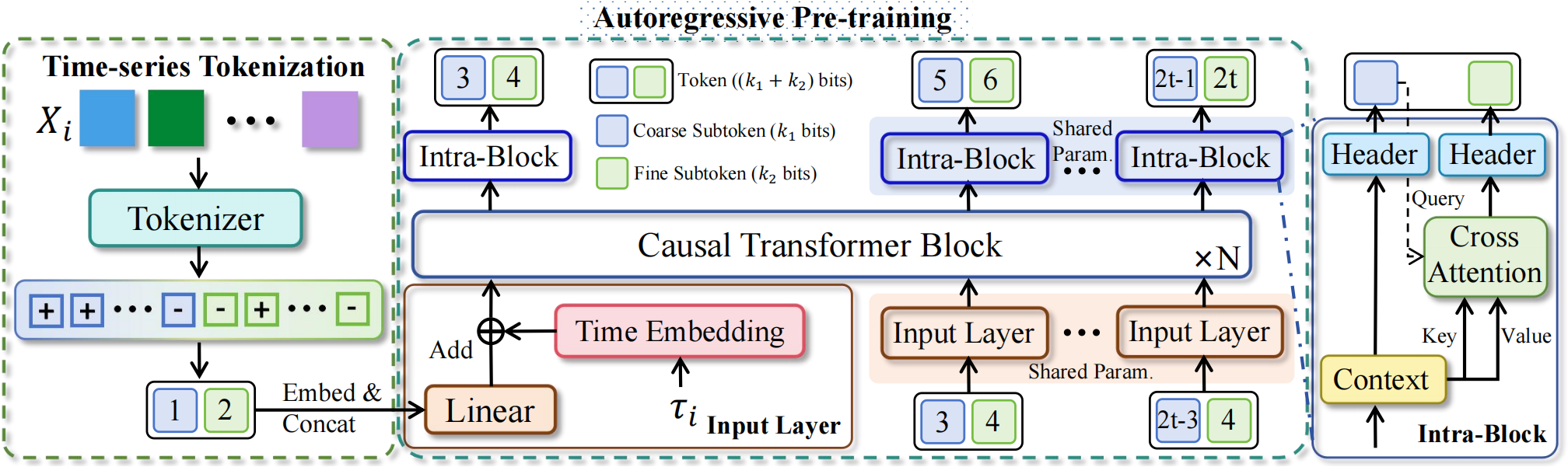} 
    \caption{The overall framework of WindFM, consisting of two main stages: (1) Time-series Tokenization, which converts continuous multivariate series into discrete hierarchical tokens, and (2) Hierarchical Autoregressive Pre-training, where a decoder-only Transformer learns to predict future tokens sequentially, incorporating Fourier-based temporal embeddings to capture periodic patterns.}
    \label{fig:overview}
\end{figure*}

\subsection{Problem Formulation}

We formulate wind power forecasting as a multivariate probabilistic time series forecasting problem. Let $\mathbf{x}_t \in \mathbb{R}^D$ denote the observation for a wind turbine site at time $t$, and let $\tau_t$ represent its corresponding timestamp information (e.g., hour, day, month). The observation vector includes the target variable, wind power, along with $D-1$ meteorological covariates. In this study, we set $D=6$, where $\mathbf{x}_t$ comprises wind power, wind speed, wind direction, air density, temperature, and atmospheric pressure.

Given a historical sequence of $L$ observations $\mathbf{x}_{1:L} = (\mathbf{x}_1, \ldots, \mathbf{x}_L)$, their timestamps $\tau_{1:L} = (\tau_1, \ldots, \tau_L)$, and the timestamps for a future horizon of $T$ steps $\tau_{L+1:L+T}$, the objective is to predict the future wind power output. Our generative model forecasts the entire multivariate vector, $\hat{\mathbf{x}}_{L+1:L+T} = (\hat{\mathbf{x}}_{L+1}, \ldots, \hat{\mathbf{x}}_{L+T})$, from which the power predictions are extracted.

Following recent generative modeling approaches, we discretize the continuous observations. A tokenizer first quantizes each multivariate vector $\mathbf{x}_t$ into a single discrete token $b_t$ from a codebook $\mathcal{C}$. Consequently, the continuous-valued sequence $\mathbf{x}_{1:L}$ is transformed into a discrete token sequence $\mathbf{b}_{1:L} = (b_1, \dots, b_L)$. The forecasting task is thereby reframed as predicting the future token sequence by maximizing its conditional likelihood, given the past tokens and all available timestamp information:
\begin{equation}
\label{eq:ts_ar}
\begin{split}
    & p(\mathbf{b}_{L+1:L+T} \mid \mathbf{b}_{1:L}, \tau_{1:L+T}) \\
    & \qquad = \prod_{t=1}^{T} p\bigl(b_{L+t} \mid \mathbf{b}_{1:L+t-1}, \tau_{1:L+T}\bigr).
\end{split}
\end{equation}
This discrete representation naturally accommodates probabilistic forecasting, which is essential for quantifying uncertainty in power systems.

\subsection{Data}

Our model is pre-trained on the Wind Integration National Dataset (WIND) Toolkit~\cite{draxl2015wind}, a public resource from the National Renewable Energy Laboratory (NREL). The dataset contains meteorological and turbine power data from 126,691 sites across the continental United States (Figure~\ref{fig:station_map}) for the years 2007–2013. With a 5-minute temporal resolution, the dataset results in an exceptionally large-scale pre-training corpus, providing a rich foundation for learning complex atmospheric dynamics.

As illustrated in Figure~\ref{fig:time_variation_pattern}, the data exhibits complex temporal dynamics, including strong diurnal and seasonal cycles, alongside non-linear relationships between meteorological variables (e.g., wind speed) and power output. These intricate patterns motivate the use of a flexible data-driven model capable of capturing such dependencies without explicit physical modeling.

\section{Methodology}

The Wind Power Foundation Model (WindFM) is a generative framework for modeling multivariate wind power time series. Its architecture, depicted in Figure~\ref{fig:overview}, is organized into two sequential stages: (1) Time-series Tokenization and (2) Hierarchical Autoregressive Pre-training. This \textit{discretize-and-generate} approach first converts continuous observations into discrete, hierarchical tokens. Subsequently, a decoder-only Transformer learns the temporal dynamics by autoregressively predicting these token sequences. This design allows the model to build a detailed, multi-scale representation of wind generation patterns, forming a robust foundation for downstream forecasting applications.

\subsection{Wind Time-series Tokenization}

\begin{figure}[t]
    \centering
    \includegraphics[width=1.0\columnwidth]{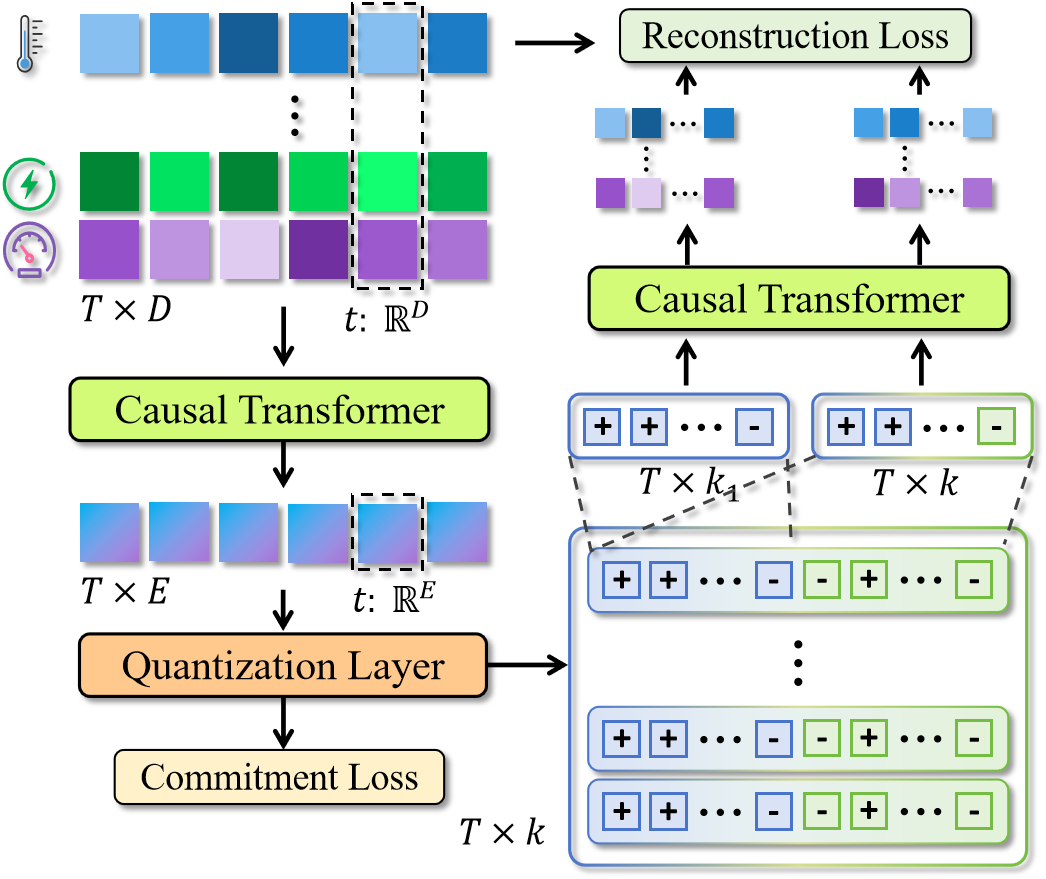}
    \caption{Architecture of the Time-series Tokenizer. It utilizes a Transformer-based autoencoder with a Binary Spherical Quantization (BSQ) layer. A composite loss objective ($\mathcal{L}_{\text{coarse}}, \mathcal{L}_{\text{fine}}, \mathcal{L}_{\text{quant}}$) enforces a coarse-to-fine representational structure within each discrete token.}
    \label{fig:tokenizer}
\end{figure}

The initial stage of WindFM transforms a continuous, $D$-dimensional time series $\mathbf{x} = (\mathbf{x}_{1}, \ldots, \mathbf{x}_{T})$, where each $\mathbf{x}_{t}\in\mathbb{R}^{D}$ represents observations at time $t$, into a corresponding sequence of discrete tokens. This is accomplished using a Transformer-based autoencoder (Figure~\ref{fig:tokenizer}), comprising an encoder $E_{\text{enc}}$, a quantizer $Q$, and a decoder $E_{\text{dec}}$.

For the quantization layer we employ Binary Spherical Quantization (BSQ)~\cite{zhao2024image}, a variant of Look-up Free Quantization (LFQ)~\cite{yu2023language}. Specifically, the encoder $E_{\text{enc}}$ first maps each input slice to a continuous latent vector $\bm{\xi}_t$. BSQ then maps this vector to a $k$-bit binary code $b_t \in \{-1,1\}^k$. A large bit length $k$ increases representational capacity but also results in an exponentially large vocabulary size ($2^k$), which poses challenges for the subsequent autoregressive model. To address this, we factorize the $k$-bit code for each time step into a coarse subtoken $b_{t}^{c}$ with bit length $k_1$ and a fine subtoken $b_{t}^{f}$ with bit length $k_2$, such that the total bit length is $k = k_1 + k_2$. The complete token $b_t$ is the concatenation of these components:
\[
  b_{t} = \bigl[b_{t}^{c},\,b_{t}^{f}\bigr],
\]
where $b_{t}^{c} \in \{-1,1\}^{k_1}$ and $b_{t}^{f} \in \{-1,1\}^{k_2}$. This decomposition replaces a single prediction over a large vocabulary of size $2^k$ with two sequential predictions over smaller, more manageable vocabularies of sizes $2^{k_1}$ and $2^{k_2}$. In our implementation, we set $k_1=k_2=10$, resulting in a total bit length of $k=20$. This choice transforms an intractable vocabulary of size $2^{20}$ into two sequential predictions over manageable vocabularies of size $2^{10}$.

To instill a coarse-to-fine structure within the tokens, the tokenizer is trained using a composite objective function that combines a hierarchical reconstruction loss with a BSQ commitment loss:
\begin{equation}
  \mathcal{L}_{\text{tokenizer}}
  = \mathcal{L}_{\text{coarse}} + \mathcal{L}_{\text{fine}} + \lambda \mathcal{L}_{\text{quant}},
\end{equation}
where $\lambda$ is a balancing hyperparameter. The loss components are defined as:
\begin{itemize}
    \item \(\mathcal{L}_{\text{coarse}} = \E\bigl[\|\mathbf{x} - E_{\text{dec}}(\mathbf{b}^{c})\|^{2}\bigr]\): This term trains the coarse subtoken \(\mathbf{b}^{c}\) to reconstruct a low-fidelity version of the input, capturing its principal structural features.
    \item \(\mathcal{L}_{\text{fine}} = \E\bigl[\|\mathbf{x} - E_{\text{dec}}(\mathbf{b})\|^{2}\bigr]\): This term evaluates the full reconstruction using the complete token \(\mathbf{b}\), guiding the fine subtoken to model the residual information needed for a more precise approximation.
    \item \(\mathcal{L}_{\text{quant}}\) is the BSQ quantization loss~\cite{zhao2024image}, which penalizes the 
    \(\ell_2\) distance between the continuous latent vectors $\bm{\xi}$ and their quantized binary codes $\mathbf{b}$. This term regularizes the encoder's output space to align with the discrete codebook.
\end{itemize}
This hierarchical objective is a key element of the tokenization process. By first training the coarse subtokens to capture dominant patterns and then training the fine subtokens to encode residual details, the desired coarse-to-fine information hierarchy is explicitly enforced, creating a structured foundation for the autoregressive modeling stage.

\subsection{Hierarchical Autoregressive Modeling}
\label{sec:autoregressive}

Following tokenization, the discrete wind feature sequences are modeled by a decoder-only Transformer, denoted as $E_{\text{ar}}$. This model employs a causal attention mechanism to ensure predictions at each time step are conditioned only on past information. The objective is to learn the joint probability distribution of the token sequence $\mathbf{b} = \{b_1, \dots, b_T\}$, which can be factorized autoregressively:
\begin{equation}
    p(\mathbf{b}) = \prod_{t=1}^{T} p(b_t | \mathbf{b}_{<t}),
\end{equation}
where $\mathbf{b}_{<t}$ denotes all tokens preceding time $t$.

The hierarchical structure of the tokens ($b_t = [b_t^{c}, b_t^{f}]$) allows for a further decomposition of the conditional probability using the chain rule, explicitly modeling the coarse-to-fine dependency:
\begin{equation}
\label{eq:chain_rule}
p(b_t | \mathbf{b}_{<t}) = p(b_t^{c} | \mathbf{b}_{<t}) \cdot p(b_t^{f} | \mathbf{b}_{<t}, b_t^{c}).
\end{equation}
This formulation enables the model to first predict the coarse subtoken, establishing a general context, before generating the fine-grained subtoken to refine the prediction.

\subsubsection{Input Representation with Temporal Encoding}

The input to the autoregressive Transformer at each time step $t$ is a vector $\mathbf{v}_t$ that combines information from the preceding token and the current timestamp, as shown in Figure~\ref{fig:overview}. This composite embedding is designed to capture both the learned discrete state of the system and its explicit temporal context.

The representation for the preceding token, $b_{t-1} = [b_{t-1}^c, b_{t-1}^f]$, is formed from its hierarchical components. The coarse and fine subtokens are projected into dense vectors via separate embedding layers, $e_c(\cdot)$ and $e_f(\cdot)$. These embeddings are then concatenated and passed through a linear projection $W_{\text{fuse}}$ to create a unified token embedding.

An important consideration in wind power forecasting is the modeling of pronounced periodic patterns, such as diurnal and seasonal cycles. To this end, a temporal encoding scheme based on Fourier features is incorporated. The timestamp for time $t$ is decomposed into multiple components (e.g., minute, hour, day, month), each normalized to $[0, 1]$. Each component $v$ is then mapped to a high-dimensional vector by projection onto a basis of sinusoidal functions:
\begin{equation}
\begin{split}
    \mathbf{\phi}(v) = [ & \sin(2\pi f_1 v), \cos(2\pi f_1 v), \ldots, \\
                        & \sin(2\pi f_N v), \cos(2\pi f_N v)]^{\top},
\end{split}
\end{equation}
where $\{f_j\}_{j=1}^N$ is a set of fixed frequencies. The resulting feature vectors for all temporal components are processed through separate linear layers and then summed to produce the final temporal embedding $\mathbf{te}_t$.

The final input vector $\mathbf{v}_t$ for the Transformer is the sum of the fused token embedding and the temporal embedding:
\begin{equation}
\label{eq:embedding}
    \mathbf{v}_t = W_{\text{fuse}}([e_c(b_{t-1}^{c}); e_f(b_{t-1}^{f})]) + \mathbf{te}_t,
\end{equation}
where $[\cdot;\cdot]$ denotes concatenation. This construction conditions the model on both the discrete state transition from the previous step and the absolute position in time, enabling it to learn and extrapolate complex temporal dynamics.

\begin{table*}[!t]
\centering
\caption{Full Results of In-Domain Forecasting Experiments. The best and second-best results are marked in \textbf{bold} and with an \underline{underline}, respectively. T denotes the forecast horizon.}
\label{tab:wind_power_full_results}
\resizebox{\textwidth}{!}{%
\begin{tabular}{lc|cc|cc|cc|cc|cc|cc|cc|cc|cc}
\toprule
\multicolumn{2}{c|}{\textbf{Setting}} & \multicolumn{2}{c|}{\textbf{WindFM (Ours)}} & \multicolumn{2}{c|}{\textbf{TimeMixer}} & \multicolumn{2}{c|}{\textbf{DLinear}} & \multicolumn{2}{c|}{\textbf{TimeXer}} & \multicolumn{2}{c|}{\textbf{iTransformer}} & \multicolumn{2}{c|}{\textbf{NSTransformer}} & \multicolumn{2}{c|}{\textbf{PatchTST}} & \multicolumn{2}{c|}{\textbf{TimesNet}} & \multicolumn{2}{c}{\textbf{FEDformer}} \\
\cmidrule(lr){3-4} \cmidrule(lr){5-6} \cmidrule(lr){7-8} \cmidrule(lr){9-10} \cmidrule(lr){11-12} \cmidrule(lr){13-14} \cmidrule(lr){15-16} \cmidrule(lr){17-18} \cmidrule(lr){19-20}
Freq. & T & MAE & RMSE & MAE & RMSE & MAE & RMSE & MAE & RMSE & MAE & RMSE & MAE & RMSE & MAE & RMSE & MAE & RMSE & MAE & RMSE \\
\midrule
\multirow{4}{*}{15min} 
& 48 & \best{0.994} & \best{1.577} & 1.198 & 1.681 & \second{1.109} & 1.703 & 1.256 & 1.665 & 1.170 & \second{1.652} & 1.197 & 1.663 & 1.204 & 1.656 & 1.196 & 1.680 & 1.246 & 1.702 \\
& 96 & \best{0.897} & \best{1.493} & 1.136 & 1.603 & \second{1.043} & 1.593 & 1.177 & 1.531 & 1.151 & \second{1.554} & 1.286 & 1.660 & 1.198 & 1.568 & 1.103 & 1.564 & 1.193 & 1.612 \\
& 144 & \best{0.879} & \best{1.497} & 1.101 & 1.528 & \second{1.045} & 1.548 & 1.135 & \second{1.479} & 1.091 & 1.492 & 1.142 & 1.568 & 1.176 & 1.520 & 1.159 & 1.615 & 1.117 & 1.544 \\
\rowcolor{lightgray} & \textbf{AVG} & \best{0.923} & \best{1.522} & 1.145 & 1.604 & \second{1.066} & 1.615 & 1.189 & \second{1.558} & 1.137 & 1.566 & 1.208 & 1.630 & 1.193 & 1.581 & 1.153 & 1.620 & 1.185 & 1.619 \\
\midrule
\multirow{4}{*}{45min} 
& 32 & \best{0.979} & \best{1.515} & 1.157 & 1.607 & \second{1.057} & 1.607 & 1.212 & 1.599 & 1.119 & \second{1.537} & 1.197 & 1.589 & 1.173 & 1.552 & 1.126 & 1.555 & 1.123 & \second{1.552} \\
& 64 & \best{0.900} & \best{1.458} & 1.074 & 1.503 & \second{1.003} & 1.484 & 1.103 & 1.425 & 1.071 & 1.466 & 1.178 & 1.506 & 1.124 & 1.474 & 1.047 & \second{1.463} & 1.159 & 1.556 \\
& 96 & \best{0.874} & \best{1.439} & 1.031 & 1.435 & \second{0.966} & 1.415 & 0.999 & \second{1.316} & 0.992 & 1.393 & 1.048 & 1.409 & 1.093 & 1.414 & 1.028 & 1.412 & 1.060 & 1.439 \\
\rowcolor{lightgray} & \textbf{AVG} & \best{0.918} & 1.471 & 1.087 & 1.515 & \second{1.009} & 1.502 & 1.105 & \best{1.447} & 1.061 & \second{1.465} & 1.141 & 1.501 & 1.130 & 1.480 & 1.067 & 1.477 & 1.114 & 1.516 \\
\midrule
\multirow{4}{*}{1H} 
& 24 & \best{0.975} & \best{1.548} & 1.241 & 1.695 & \second{1.077} & 1.656 & 1.161 & \second{1.532} & 1.160 & 1.607 & 1.239 & 1.646 & 1.208 & 1.614 & 1.146 & 1.604 & 1.174 & 1.626 \\
& 48 & \best{0.906} & \best{1.479} & 1.084 & 1.502 & \second{1.013} & 1.513 & 1.100 & \second{1.453} & 1.058 & 1.481 & 1.111 & 1.498 & 1.150 & 1.506 & 1.081 & 1.496 & 1.133 & 1.516 \\
& 72 & \best{0.864} & 1.418 & 1.032 & 1.459 & \second{0.969} & 1.427 & 1.063 & \best{1.400} & 1.017 & 1.414 & 1.075 & 1.438 & 1.058 & \second{1.410} & 1.076 & 1.455 & 1.112 & 1.483 \\
\rowcolor{lightgray} & \textbf{AVG} & \best{0.915} & \second{1.482} & 1.119 & 1.552 & \second{1.020} & 1.532 & 1.108 & \best{1.462} & 1.078 & 1.501 & 1.142 & 1.527 & 1.139 & 1.510 & 1.101 & 1.518 & 1.140 & 1.542 \\
\midrule
\multirow{4}{*}{2H} 
& 16 & \best{1.017} & \best{1.549} & 1.216 & 1.646 & \second{1.127} & 1.656 & 1.168 & \second{1.575} & 1.157 & 1.600 & 1.233 & 1.630 & 1.203 & 1.614 & 1.139 & 1.585 & 1.194 & 1.618 \\
& 32 & \best{0.889} & \best{1.408} & \second{1.000} & \second{1.415} & 1.006 & 1.454 & 1.115 & 1.473 & 1.041 & 1.434 & 1.106 & 1.447 & 1.105 & 1.441 & 1.053 & 1.434 & 1.037 & 1.429 \\
& 64 & \best{0.847} & 1.381 & 0.961 & \second{1.347} & \second{0.936} & 1.351 & 1.068 & 1.412 & 0.963 & 1.351 & 1.037 & 1.352 & 1.010 & \best{1.327} & 0.989 & 1.354 & 0.972 & 1.366 \\
\rowcolor{lightgray} & \textbf{AVG} & \best{0.918} & \best{1.446} & 1.059 & 1.469 & \second{1.023} & 1.487 & 1.117 & 1.487 & 1.054 & 1.462 & 1.125 & 1.476 & 1.106 & 1.461 & 1.060 & \second{1.458} & 1.068 & 1.471 \\
\midrule
\rowcolor{mediumgray}
\multicolumn{2}{c|}{\textbf{Average}} & \best{0.918} & \best{1.480} & 1.103 & 1.535 & \second{1.029} & 1.534 & 1.130 & \second{1.488} & 1.083 & 1.498 & 1.154 & 1.534 & 1.142 & 1.508 & 1.095 & 1.518 & 1.127 & 1.537 \\
\rowcolor{lightgray}
\multicolumn{2}{c|}{\textbf{1\textsuperscript{st} Count}} & \multicolumn{2}{c|}{\textbf{30}} & \multicolumn{2}{c|}{0} & \multicolumn{2}{c|}{0} & \multicolumn{2}{c|}{3} & \multicolumn{2}{c|}{0} & \multicolumn{2}{c|}{0} & \multicolumn{2}{c|}{1} & \multicolumn{2}{c|}{0} & \multicolumn{2}{c}{0} \\
\bottomrule
\end{tabular}%
}
\end{table*}

\begin{table*}[!t]
\centering
\caption{Full Results of Zero-Shot Forecasting Experiments. The best and second-best results are marked in \textbf{bold} and with an \underline{underline}, respectively. T denotes the forecast horizon.}
\label{tab:zeroshot_full_results}
\resizebox{\textwidth}{!}{%
\begin{tabular}{lc|cc|cc|cc|cc|cc|cc|cc|cc|cc|cc|cc|cc|cc}
\toprule
\multicolumn{2}{c|}{\textbf{Setting}} & \multicolumn{2}{c|}{\textbf{WindFM (Ours)}} & \multicolumn{2}{c|}{\textbf{$\text{TimeMOE}_{small}$}} & \multicolumn{2}{c|}{\textbf{$\text{TimeMOE}_{base}$}} & \multicolumn{2}{c|}{\textbf{$\text{Moirai}_{small}$}} & \multicolumn{2}{c|}{\textbf{$\text{Moirai}_{base}$}} & \multicolumn{2}{c|}{\textbf{$\text{Moirai}_{large}$}} & \multicolumn{2}{c|}{\textbf{TimesFM}} & \multicolumn{2}{c|}{\textbf{$\text{Moment}_{small}$}} & \multicolumn{2}{c|}{\textbf{$\text{Moment}_{base}$}} & \multicolumn{2}{c|}{\textbf{$\text{Moment}_{large}$}} & \multicolumn{2}{c|}{\textbf{$\text{Chronos}_{small}$}} & \multicolumn{2}{c|}{\textbf{$\text{Chronos}_{base}$}} & \multicolumn{2}{c}{\textbf{$\text{Chronos}_{large}$}} \\
\cmidrule(lr){3-4} \cmidrule(lr){5-6} \cmidrule(lr){7-8} \cmidrule(lr){9-10} \cmidrule(lr){11-12} \cmidrule(lr){13-14} \cmidrule(lr){15-16} \cmidrule(lr){17-18} \cmidrule(lr){19-20} \cmidrule(lr){21-22} \cmidrule(lr){23-24} \cmidrule(lr){25-26} \cmidrule(lr){27-28}
Freq. & T & MAE & RMSE & MAE & RMSE & MAE & RMSE & MAE & RMSE & MAE & RMSE & MAE & RMSE & MAE & RMSE & MAE & RMSE & MAE & RMSE & MAE & RMSE & MAE & RMSE & MAE & RMSE & MAE & RMSE \\
\midrule
\multirow{4}{*}{15min} 
& 48 & \best{0.994} & \best{1.577} & 1.059 & 1.667 & \second{1.058} & \second{1.638} & 1.248 & 1.769 & 1.249 & 1.769 & 1.250 & 1.770 & 1.249 & 2.098 & 1.212 & 1.773 & 1.212 & 1.773 & 1.209 & 1.772 & 1.157 & 2.006 & 1.155 & 2.049 & 1.151 & 2.004 \\
& 96 & \best{0.897} & \best{1.493} & 0.993 & 1.539 & \second{0.991} & \second{1.514} & 1.581 & 1.951 & 1.362 & 1.772 & 1.381 & 1.785 & 1.081 & 1.691 & 1.101 & 1.600 & 1.096 & 1.599 & 1.097 & 1.597 & 1.060 & 1.876 & 1.060 & 1.890 & 1.072 & 1.906 \\
& 144 & \best{0.879} & \second{1.497} & 0.981 & 1.511 & \second{0.979} & \best{1.487} & 1.615 & 1.960 & 1.340 & 1.735 & 1.338 & 1.728 & 1.023 & 1.587 & 1.071 & 1.553 & 1.066 & 1.551 & 1.689 & 1.549 & 1.041 & 1.787 & 1.043 & 1.800 & 1.076 & 1.842 \\
\rowcolor{lightgray} & \textbf{AVG} & \best{0.923} & \best{1.522} & 1.011 & 1.572 & \second{1.009} & \second{1.546} & 1.481 & 1.893 & 1.317 & 1.759 & 1.323 & 1.761 & 1.118 & 1.792 & 1.128 & 1.642 & 1.125 & 1.641 & 1.332 & 1.639 & 1.086 & 1.890 & 1.086 & 1.913 & 1.100 & 1.917 \\
\midrule
\multirow{4}{*}{45min} 
& 32 & \best{0.979} & \best{1.515} & \second{1.035} & 1.572 & 1.038 & \second{1.546} & 1.176 & 1.627 & 1.176 & 1.627 & 1.177 & 1.627 & 1.079 & 1.716 & 1.123 & 1.615 & 1.120 & 1.614 & 1.119 & 1.614 & 1.100 & 1.897 & 1.099 & 1.864 & 1.077 & 1.796 \\
& 64 & \best{0.900} & \best{1.458} & 0.981 & 1.480 & 0.981 & \second{1.463} & 1.116 & 1.532 & 1.116 & 1.532 & 1.115 & 1.532 & 1.040 & 1.648 & 1.060 & 1.520 & 1.058 & 1.518 & 1.056 & 1.517 & 1.051 & 1.810 & 1.050 & 1.794 & \second{1.039} & 1.754 \\
& 96 & \best{0.874} & 1.439 & 0.942 & \second{1.412} & \second{0.941} & \best{1.395} & 1.486 & 1.800 & 1.248 & 1.606 & 1.269 & 1.622 & 0.963 & 1.519 & 1.009 & 1.442 & 1.004 & 1.441 & 1.005 & 1.439 & 0.951 & 1.684 & 0.957 & 1.674 & 0.958 & 1.669 \\
\rowcolor{lightgray} & \textbf{AVG} & \best{0.918} & \second{1.471} & 0.986 & 1.488 & 0.987 & \best{1.468} & 1.259 & 1.653 & 1.180 & 1.588 & 1.187 & 1.594 & 1.027 & 1.628 & 1.064 & 1.526 & 1.061 & 1.524 & 1.060 & 1.523 & 1.034 & 1.797 & 1.035 & 1.777 & \second{1.025} & 1.740 \\
\midrule
\multirow{4}{*}{1H} 
& 24 & \best{0.975} & \best{1.548} & 1.065 & 1.639 & 1.065 & \second{1.613} & 1.201 & 1.682 & 1.200 & 1.682 & 1.200 & 1.682 & 1.139 & 1.833 & 1.146 & 1.676 & 1.147 & 1.676 & 1.145 & 1.675 & 1.045 & 1.719 & 1.051 & 1.715 & \second{1.040} & 1.708 \\
& 48 & \best{0.906} & \second{1.479} & \second{0.982} & 1.496 & 0.984 & \best{1.477} & 1.120 & 1.544 & 1.121 & 1.545 & 1.122 & 1.545 & 1.064 & 1.741 & 1.063 & 1.536 & 1.063 & 1.536 & 1.060 & 1.535 & 1.007 & 1.689 & 1.004 & 1.675 & 1.011 & 1.702 \\
& 72 & \best{0.864} & \second{1.418} & \second{0.943} & 1.430 & 0.944 & \best{1.416} & 1.073 & 1.474 & 1.073 & 1.473 & 1.096 & 1.473 & 0.998 & 1.558 & 1.018 & 1.466 & 1.015 & 1.465 & 1.015 & 1.464 & 0.969 & 1.642 & 0.971 & 1.651 & 0.975 & 1.650 \\
\rowcolor{lightgray} & \textbf{AVG} & \best{0.915} & \best{1.482} & \second{0.996} & 1.522 & 1.000 & \second{1.502} & 1.131 & 1.567 & 1.131 & 1.567 & 1.139 & 1.567 & 1.067 & 1.711 & 1.076 & 1.559 & 1.075 & 1.559 & 1.073 & 1.558 & 0.999 & 1.683 & 1.009 & 1.680 & 1.009 & 1.687 \\
\midrule
\multirow{4}{*}{2H} 
& 16 & \second{1.017} & \best{1.549} & 1.093 & 1.640 & 1.096 & 1.620 & 1.216 & 1.682 & 1.213 & 1.681 & 1.122 & 1.682 & 1.084 & 1.740 & 1.155 & 1.671 & 1.158 & 1.670 & 1.155 & 1.670 & 1.055 & 1.656 & \best{1.043} & \second{1.575} & 1.063 & 1.673 \\
& 32 & \best{0.889} & \best{1.408} & 0.976 & 1.452 & 0.978 & \second{1.437} & 1.096 & 1.488 & 1.095 & 1.487 & 1.073 & 1.449 & \second{0.972} & 1.557 & 1.036 & 1.472 & 1.033 & 1.471 & 1.032 & 1.471 & 0.978 & 1.560 & 0.983 & 1.553 & 0.980 & 1.546 \\
& 64 & \best{0.847} & 1.381 & 0.920 & \second{1.356} & 0.923 & \best{1.347} & 1.018 & 1.383 & 1.021 & 1.384 & 1.216 & 1.383 & 0.923 & 1.478 & 0.962 & 1.369 & 0.960 & 1.367 & 0.958 & 1.367 & \second{0.903} & 1.463 & 0.910 & 1.459 & 0.910 & 1.465 \\
\rowcolor{lightgray} & \textbf{AVG} & \best{0.918} & \best{1.463} & 0.996 & 1.483 & 0.999 & \second{1.468} & 1.110 & 1.518 & 1.110 & 1.517 & 1.137 & 1.505 & 0.993 & 1.592 & 1.051 & 1.504 & 1.050 & 1.503 & 1.048 & 1.503 & \second{0.979} & 1.560 & 0.979 & 1.529 & 0.984 & 1.561 \\
\midrule
\rowcolor{mediumgray}
\multicolumn{2}{c|}{\textbf{Average}} & \best{0.918} & \best{1.480} & \second{0.997} & 1.516 & 0.998 & \second{1.496} & 1.246 & 1.658 & 1.185 & 1.608 & 1.197 & 1.607 & 1.051 & 1.681 & 1.080 & 1.558 & 1.078 & 1.557 & 1.128 & 1.556 & 1.026 & 1.732 & 1.027 & 1.731 & 1.029 & 1.726 \\
\rowcolor{lightgray}
\multicolumn{2}{c|}{\textbf{1\textsuperscript{st} Count}} & \multicolumn{2}{c|}{\textbf{27}} & \multicolumn{2}{c|}{0} & \multicolumn{2}{c|}{6} & \multicolumn{2}{c|}{0} & \multicolumn{2}{c|}{0} & \multicolumn{2}{c|}{0} & \multicolumn{2}{c|}{0} & \multicolumn{2}{c|}{0} & \multicolumn{2}{c|}{0} & \multicolumn{2}{c|}{0} & \multicolumn{2}{c|}{0} & \multicolumn{2}{c|}{1} & \multicolumn{2}{c}{0} \\
\bottomrule
\end{tabular}%
}
\end{table*}

\subsubsection{Sequential Prediction Process}

The sequence of input vectors $\{\mathbf{v}_1, \dots, \mathbf{v}_{t}\}$ is processed by the Transformer $E_{\text{ar}}$ to produce a sequence of contextualized hidden states. The final hidden state, $\mathbf{h}_t$, encapsulates historical information up to time $t-1$ and is used to predict the token $b_t$.

\textbf{Coarse Subtoken Prediction.} The history vector $\mathbf{h}_t$ is projected by a linear head $W_c$ to compute the logits for the coarse subtoken distribution:
\begin{equation}
    p(b_{t}^{c} | \mathbf{b}_{< t}) = \text{softmax}(W_c \mathbf{h}_t).
\end{equation}

\textbf{Fine Subtoken Prediction.} To model the conditional dependency in Eq. (\ref{eq:chain_rule}), the context is updated with the predicted coarse subtoken, $\hat{b}_{t}^{c}$. During training, $\hat{b}_{t}^{c}$ is sampled from the predicted distribution $p(b_{t}^{c} | \mathbf{b}_{< t})$ instead of using the ground-truth value. This strategy helps mitigate exposure bias by better aligning the training process with inference-time generation, where ground-truth tokens are unavailable. A cross-attention mechanism updates the context, where the embedding of $\hat{b}_{t}^{c}$ serves as the query and the history vector $\mathbf{h}_t$ provides the key and value. The output is then projected by a second head $W_f$:
\begin{equation}
\begin{aligned}
    \mathbf{h}^{\text{update}}_t &= \text{CrossAttn}(q=e_c(\hat{b}_{t}^{c}), k=\mathbf{h}_t, v=\mathbf{h}_t), \\
    p(b_{t}^{f} | \mathbf{b}_{< t}, b_{t}^{c}) &= \text{softmax}(W_f \mathbf{h}^{\text{update}}_t).
\end{aligned}
\end{equation}
The overall training objective $\mathcal{L}_{\text{ar}}$ is to minimize the negative log-likelihood of the data sequence, summed over both prediction steps:
\begin{equation}
    \mathcal{L}_{\text{ar}} = - \E_{\mathbf{b} \sim \mathcal{D}} \sum_{t=1}^{T} \left[ \log p(b_t^{c} | \mathbf{b}_{<t}) + \log p(b_t^{f} | \mathbf{b}_{<t}, b_t^{c}) \right],
\end{equation}
where $\mathcal{D}$ denotes the data distribution.

\subsubsection{Model Configurations and Training Details}

The pre-training dataset is constructed from the WIND Toolkit, containing time series from 126,499 sites for the years 2007–2012. An additional 192 sites are reserved for validation. To enable the model to handle different time scales, the original 5-minute resolution data is augmented by downsampling to 15-minute, 30-minute, 1-hour, 4-hour, and 1-day resolutions. To create a balanced multi-resolution training corpus, we subsample the 5-minute and 15-minute data by factors of 30 and 5, respectively. All input sequences have a maximum length of 512 time steps. 

The tokenizer's encoder and decoder are 3-layer Transformers, and the autoregressive model is a 4-layer decoder-only Transformer. Both components use a model dimension of 256 and a feed-forward network dimension of 512. The number of attention heads is 4 for the tokenizer and 8 for the autoregressive model, which also uses a dropout rate of 0.2. For temporal encoding, we extract features for minute, hour, day of the week, day of the month, and month, with each component independently mapped to a 10-dimensional Fourier feature representation. Positional information is encoded using Rotary Position Embeddings (RoPE)~\cite{su2024roformer}. Training stability is enhanced through a Pre-Layer Normalization (Pre-LN)~\cite{xiong2020layer} architecture with Root Mean Square Layer Normalization (RMSNorm)~\cite{zhang2019root}.

The BSQ module uses a commitment weight of $\beta=0.05$, entropy penalty weights $\gamma_0=1.0$ and $\gamma=1.1$, an entropy scale $\zeta=0.05$, and a quantization group size of 5. The balancing hyperparameter $\lambda$ for the quantization loss is set to 1.0. The entire model is trained using the AdamW optimizer~\cite{loshchilov2017decoupled} with a learning rate of 5e-4 and a weight decay of 0.01.

\section{Experiment}

\subsection{Baselines}

To rigorously evaluate WindFM, we benchmark it against a comprehensive suite of state-of-the-art time series models, categorized into two distinct groups.

\begin{itemize}
    \item \textbf{Site-Specific Time Series Models:} This category includes modern models that are trained from scratch on the target wind power dataset. The selection covers a range of architectures: Transformer-based models (iTransformer~\cite{liuitransformer}, PatchTST~\cite{nie2022time}, Non-stationary Transformer (NSTransformer)~\cite{liu2022non}, FEDformer~\cite{zhou2022fedformer}, TimeXer~\cite{wang2024timexer}), MLP-based models (TimeMixer~\cite{wang2024timemixer}), CNN-based models (TimesNet~\cite{wu2022timesnet}), and a linear model (DLinear~\cite{zeng2023transformers}).

    \item \textbf{Time-series Foundation Models:} This group comprises large-scale foundation models designed for general time series analysis. We evaluate these models in a zero-shot setting. The models include TimeMOE~\cite{shi2024time}, Moirai~\cite{woo2024unified}, TimesFM~\cite{das2024decoder}, Moment~\cite{goswami2024moment}, and Chronos~\cite{ansari2024chronos}.
\end{itemize}

For probabilistic forecasting, we leverage WindFM's generative capabilities and compare it against foundation models that natively support probabilistic outputs (Moirai and Chronos). We also adapt high-performing site-specific models (iTransformer, TimeMixer, TimeXer, and DLinear) by replacing their output layers with a projection head that parameterizes a Gaussian distribution ($\mu, \sigma$) and training them with a Negative Log-Likelihood (NLL) loss.

\begin{table}[ht]
\centering
\caption{Hyperparameter configurations for the site-specific baseline models. Values for the two evaluated sets (smaller/larger) are separated by a slash.}
\label{tab:baseline_hyperparams}
\resizebox{\columnwidth}{!}{%
\begin{tabular}{l cccc}
\toprule
\textbf{Model} & \textbf{Layers} & \textbf{$\mathbf{d}_{\text{model}}$} & \textbf{$\mathbf{d}_{\text{ff}}$ (FFN Dim.)} & \textbf{Heads} \\
\midrule
TimeXer       & 3 / 5 & 128 / 256 & 256 / 512 & 4 / 8 \\
TimesNet      & 3 / 5 & 128 / 256 & 256 / 512 & ---   \\
TimeMixer     & 3 / 5 & 128 / 256 & 256 / 512 & 4 / 8 \\
PatchTST      & 3 / 5 & 128 / 256 & 256 / 512 & 4 / 8 \\
NSTransformer & 2 / 3 & 128 / 256 & 256 / 512 & 4 / 8 \\
FEDformer     & 2 / 3 & 128 / 256 & 256 / 512 & 4 / 8 \\
iTransformer  & 3 / 5 & 128 / 256 & 256 / 512 & 4 / 8 \\
\bottomrule
\end{tabular}}
\end{table}

All site-specific models are trained under a unified protocol for fair comparison. We use the Adam optimizer with a learning rate of $5 \times 10^{-4}$ and a batch size of 256. Models are trained for up to 12 epochs, with early stopping based on validation loss (patience of 3 epochs). For each model, we evaluate two hyperparameter sets (smaller/larger) and select the configuration with the best validation performance. For DLinear, we test two modes: a shared linear layer across all variates and separate layers per variate. Architectural hyperparameters are detailed in Table~\ref{tab:baseline_hyperparams}.

WindFM's autoregressive generation is controlled by temperature scaling ($T$) and top-$p$ (nucleus) sampling. Final point forecasts are the mean of multiple sampled trajectories. Inference hyperparameters are tuned for each metric: for MAE, we use $T=0.6$, top-$p=0.9$, and 20 samples; for RMSE, which is more sensitive to large errors, we use a higher temperature of $T=0.9$. For probabilistic forecasting, we generate an empirical distribution from 100 samples with $T=1.0$ and top-$p=1.0$ to maximize diversity.

\subsection{Tasks and Evaluation Metrics}

\begin{figure}[t!]
    \centering
    \includegraphics[width=1.0\columnwidth]{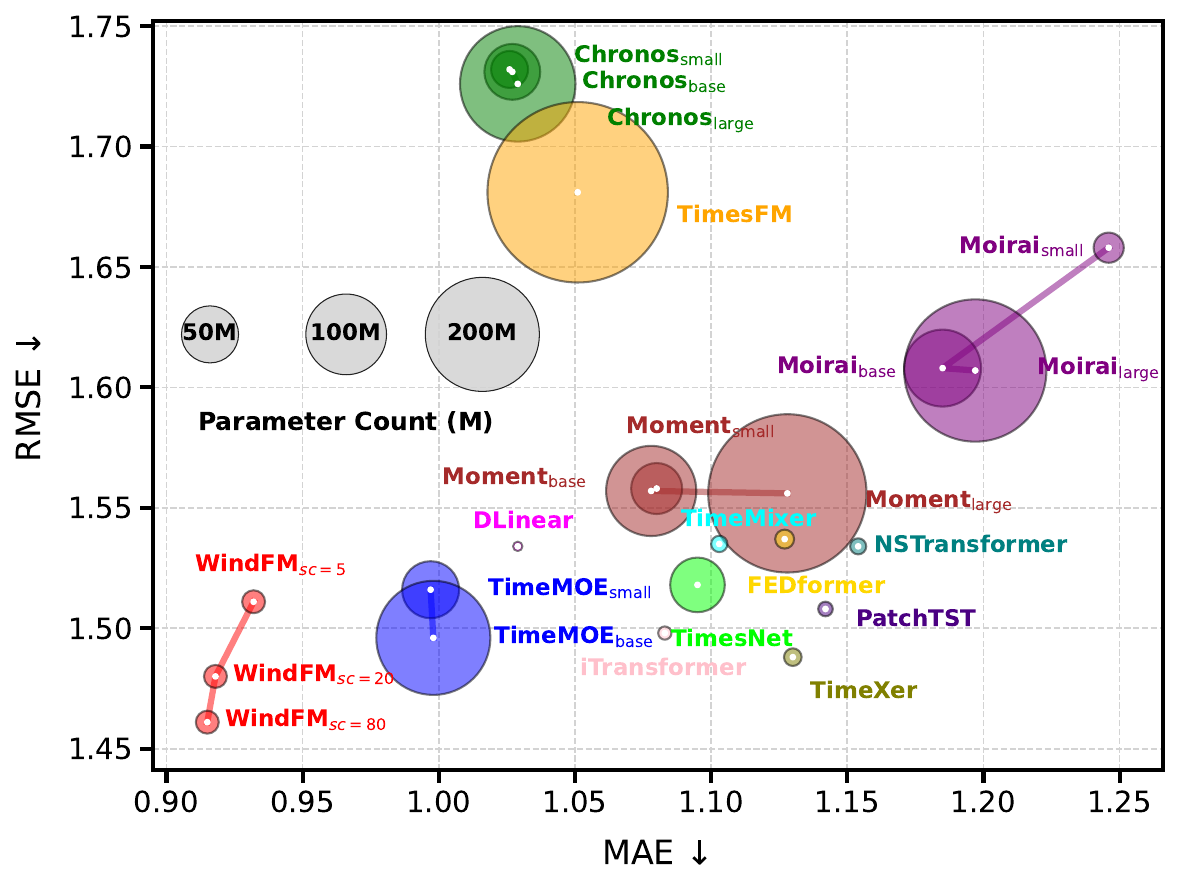}
    \caption{Zero-shot forecasting performance versus model size. WindFM (red) achieves state-of-the-art accuracy (lower MAE and RMSE) while being more parameter-efficient than other leading foundation models. The bubble size is proportional to the parameter count. \texttt{sc} denotes the sample count for WindFM during inference.
    }
    \label{fig:model_param_comparison}
\end{figure}

We evaluate WindFM on three tasks designed to assess key aspects of its performance.

\textbf{1. Zero-Shot Forecasting on Unseen Domestic Sites.} This task assesses zero-shot accuracy on unseen data from the same geographical region as the pre-training set (7 hold-out WIND Toolkit sites, year 2013). To evaluate versatility, we test on four frequencies: two seen during pre-training (15-minute, 1-hour) and two unseen (45-minute, 2-hour). The lookback window length equals the forecast horizon. We test three horizons for each frequency (15-min: 48, 96, 144; 45-min: 32, 64, 96; 1-hour: 24, 48, 72; 2-hour: 16, 32, 64). Performance is measured by Mean Absolute Error (MAE) and Root Mean Squared Error (RMSE).

\begin{table*}[t!]
\centering
\caption{Full Results of Probabilistic Forecasting Experiments. The best and second-best results are marked in \textbf{bold} and with an \underline{underline}, respectively. T denotes the forecast horizon.}
\label{tab:probabilistic_full_results}
\resizebox{\textwidth}{!}{%
\begin{tabular}{lc|cc|cc|cc|cc|cc|cc|cc|cc|cc|cc}
\toprule
\multicolumn{2}{c|}{\textbf{Setting}} & \multicolumn{2}{c|}{\textbf{WindFM (Ours)}} & \multicolumn{2}{c|}{\textbf{Moirai\_base}} & \multicolumn{2}{c|}{\textbf{Moirai\_large}} & \multicolumn{2}{c|}{\textbf{Chronos\_small}} & \multicolumn{2}{c|}{\textbf{Chronos\_base}} & \multicolumn{2}{c|}{\textbf{Chronos\_large}} & \multicolumn{2}{c|}{\textbf{TimeMixer}} & \multicolumn{2}{c|}{\textbf{TimeXer}} & \multicolumn{2}{c|}{\textbf{DLinear}} & \multicolumn{2}{c}{\textbf{iTransformer}} \\
\cmidrule(lr){3-4} \cmidrule(lr){5-6} \cmidrule(lr){7-8} \cmidrule(lr){9-10} \cmidrule(lr){11-12} \cmidrule(lr){13-14} \cmidrule(lr){15-16} \cmidrule(lr){17-18} \cmidrule(lr){19-20} \cmidrule(lr){21-22}
Freq. & T & CRPS~$\downarrow$ & AQL~$\downarrow$ & CRPS & AQL & CRPS & AQL & CRPS & AQL & CRPS & AQL & CRPS & AQL & CRPS & AQL & CRPS & AQL & CRPS & AQL & CRPS & AQL \\
\midrule
\multirow{4}{*}{15min} 
& 48 & \best{0.777} & \best{0.418} & 1.043 & 0.565 & 1.043 & 0.565 & 0.886 & 0.479 & 0.884 & 0.478 & \second{0.881} & \second{0.477} & 1.752 & 0.876 & 1.747 & 0.873 & 2.097 & 1.049 & 6.236 & 3.117 \\
& 96 & \best{0.707} & \best{0.381} & 0.971 & 0.525 & 1.007 & 0.536 & \second{0.879} & \second{0.465} & 0.880 & 0.466 & 0.894 & 0.472 & 1.618 & 0.809 & 1.618 & 0.809 & 1.971 & 0.985 & 1.897 & 0.948 \\
& 144 & \best{0.693} & \best{0.374} & 0.968 & 0.522 & 0.986 & 0.526 & \second{0.880} & \second{0.463} & 0.886 & 0.466 & 0.913 & 0.480 & 1.714 & 0.857 & 1.526 & 0.763 & 1.863 & 0.931 & 1.777 & 0.888 \\
\rowcolor{lightgray} & \textbf{AVG} & \best{0.726} & \best{0.391} & 0.994 & 0.537 & 1.012 & 0.542 & \second{0.882} & \second{0.469} & 0.883 & 0.470 & 0.896 & 0.476 & 1.695 & 0.847 & 1.630 & 0.815 & 1.977 & 0.988 & 3.303 & 1.651 \\
\midrule
\multirow{4}{*}{45min} 
& 32 & \best{0.746} & \best{0.403} & 0.983 & 0.534 & 0.983 & 0.534 & 0.840 & 0.456 & 0.837 & 0.454 & \second{0.816} & \second{0.443} & 1.732 & 0.866 & 1.617 & 0.808 & 1.862 & 0.931 & 3.664 & 1.832 \\
& 64 & \best{0.687} & \best{0.372} & 0.924 & 0.504 & 0.924 & 0.503 & 0.793 & 0.430 & 0.791 & 0.429 & \second{0.782} & \second{0.424} & 1.627 & 0.814 & 1.522 & 0.761 & 1.703 & 0.851 & 2.069 & 1.034 \\
& 96 & \best{0.652} & \best{0.354} & 0.893 & 0.483 & 0.923 & 0.496 & \second{0.774} & \second{0.412} & 0.775 & 0.412 & 0.777 & 0.414 & 1.558 & 0.779 & 1.383 & 0.691 & 1.621 & 0.811 & 1.817 & 0.908 \\
\rowcolor{lightgray} & \textbf{AVG} & \best{0.695} & \best{0.376} & 0.933 & 0.507 & 0.943 & 0.511 & 0.802 & 0.433 & 0.801 & 0.432 & \second{0.792} & \second{0.427} & 1.639 & 0.820 & 1.507 & 0.753 & 1.729 & 0.864 & 2.517 & 1.258 \\
\midrule
\multirow{4}{*}{1H} 
& 24 & \best{0.757} & \best{0.408} & 0.999 & 0.542 & 0.999 & 0.543 & 0.809 & 0.438 & 0.812 & 0.439 & \second{0.800} & \second{0.434} & 1.698 & 0.850 & 1.689 & 0.845 & 1.720 & 0.860 & 2.445 & 1.224 \\
& 48 & \best{0.693} & \best{0.375} & 0.926 & 0.505 & 0.926 & 0.505 & 0.764 & 0.414 & \second{0.757} & \second{0.411} & 0.762 & 0.413 & 1.542 & 0.771 & 1.485 & 0.743 & 1.576 & 0.788 & 2.003 & 1.002 \\
& 72 & \best{0.657} & \best{0.356} & 0.880 & 0.481 & 0.880 & 0.480 & \second{0.753} & \second{0.405} & 0.754 & 0.406 & 0.758 & 0.408 & 1.506 & 0.753 & 1.415 & 0.708 & 1.458 & 0.729 & 1.688 & 0.844 \\
\rowcolor{lightgray} & \textbf{AVG} & \best{0.702} & \best{0.380} & 0.935 & 0.509 & 0.935 & 0.509 & 0.775 & 0.419 & \second{0.774} & 0.419 & 0.773 & \second{0.418} & 1.582 & 0.791 & 1.530 & 0.765 & 1.585 & 0.792 & 2.045 & 1.023 \\
\midrule
\multirow{4}{*}{2H} 
& 16 & \best{0.770} & \best{0.415} & 1.014 & 0.551 & 1.014 & 0.551 & 0.790 & 0.428 & \second{0.783} & \second{0.425} & 0.796 & 0.432 & 1.685 & 0.842 & 1.679 & 0.840 & 1.904 & 0.952 & 2.474 & 1.237 \\
& 32 & \best{0.668} & \best{0.362} & 0.904 & 0.494 & 0.904 & 0.494 & 0.740 & 0.402 & 0.741 & 0.403 & \second{0.735} & \second{0.400} & 1.627 & 0.814 & 1.469 & 0.734 & 1.731 & 0.866 & 1.962 & 0.981 \\
& 64 & \best{0.628} & \best{0.341} & 0.832 & 0.455 & 0.831 & 0.455 & \second{0.679} & \second{0.368} & 0.680 & 0.370 & 0.681 & 0.370 & 1.583 & 0.791 & 1.356 & 0.678 & 1.646 & 0.823 & 1.537 & 0.768 \\
\rowcolor{lightgray} & \textbf{AVG} & \best{0.689} & \best{0.373} & 0.917 & 0.500 & 0.916 & 0.500 & 0.736 & 0.399 & 0.735 & 0.399 & \second{0.737} & \second{0.401} & 1.632 & 0.816 & 1.501 & 0.751 & 1.760 & 0.880 & 1.991 & 0.995 \\
\midrule
\rowcolor{mediumgray}
\multicolumn{2}{c|}{\textbf{Average}} & \best{0.703} & \best{0.380} & 0.945 & 0.513 & 0.952 & 0.516 & 0.799 & 0.430 & \second{0.798} & \second{0.429} & 0.799 & 0.430 & 1.637 & 0.818 & 1.542 & 0.771 & 1.763 & 0.948 & 2.464 & 1.232 \\
\rowcolor{lightgray}
\multicolumn{2}{c|}{\textbf{1\textsuperscript{st} Count}} & \multicolumn{2}{c|}{\textbf{34}} & \multicolumn{2}{c|}{\textbf{0}} & \multicolumn{2}{c|}{\textbf{0}} & \multicolumn{2}{c|}{\textbf{0}} & \multicolumn{2}{c|}{\textbf{0}} & \multicolumn{2}{c|}{\textbf{0}} & \multicolumn{2}{c|}{\textbf{0}} & \multicolumn{2}{c|}{\textbf{0}} & \multicolumn{2}{c|}{\textbf{0}} & \multicolumn{2}{c}{\textbf{0}} \\
\bottomrule
\end{tabular}%
}
\end{table*}

\textbf{2. Probabilistic Forecasting.} This task evaluates the quality of WindFM's probabilistic forecasts. The setup, including datasets and horizons, mirrors the deterministic forecasting task. We use two standard metrics: Continuous Ranked Probability Score (CRPS) and Average Quantile Loss (AQL) \cite{abedinia2024wind, zhu2025novel}.
The CRPS is defined as:
\begin{equation}
    \text{CRPS}(F, y) = \int_{-\infty}^{\infty} (F(z) - \mathbf{1}\{y \le z\})^2 dz,
\end{equation}
where $F$ is the predicted cumulative distribution function (CDF) and $y$ is the ground-truth observation. The AQL, also known as the pinball loss, is calculated by averaging the quantile loss $\rho_q$ over a set of quantiles $\mathcal{Q}$:
\begin{equation}
    \rho_q(y, \hat{y}_q) = 
    \begin{cases} 
        q(y - \hat{y}_q) & \text{if } y \ge \hat{y}_q \\
        (1-q)(\hat{y}_q - y) & \text{if } y < \hat{y}_q,
    \end{cases}
\end{equation}
where $\hat{y}_q$ is the predicted $q$-th quantile. A lower value indicates a better forecast for both CRPS and AQL.

\begin{table}[!t]
\centering
\caption{Cross-Geography Generalization: Zero-Shot Forecasting on a Chinese Dataset. The best and second-best results are marked in \textbf{bold} and with an \underline{underline}, respectively.}
\label{tab:my_experiment_results}
\resizebox{\columnwidth}{!}{%
\begin{tabular}{c|cc|cc|cc}
\toprule
\multicolumn{1}{c|}{\multirow{2}{*}{\textbf{Model}}} & \multicolumn{2}{c|}{\textbf{T=48}} & \multicolumn{2}{c|}{\textbf{T=96}} & \multicolumn{2}{c}{\textbf{T=144}} \\
\cmidrule(lr){2-3} \cmidrule(lr){4-5} \cmidrule(lr){6-7}
\multicolumn{1}{c|}{} & MAE & RMSE & MAE & RMSE & MAE & RMSE \\
\midrule

TimeMixer       & 1.487        & 2.007          & 1.379        & 1.893          & 1.410        & 1.867 \\
DLinear         & 1.442        & 1.996          & 1.423        & 1.954          & 1.534        & 2.003 \\
TimeXer         & \second{1.435} & \best{1.826}   & 1.440        & \second{1.810}          & 1.549        & 1.907 \\
iTransformer    & 1.456        & 1.997          & 1.492        & 2.001          & 1.564        & 2.035 \\
NSTransformer   & 1.555        & 1.956          & 1.475        & 1.831          & 1.423        & 1.733 \\
PatchTST        & 1.436        & \second{1.887} & 1.383        & 1.816          & 1.524        & 1.936 \\
TimesNet        & 1.477        & 1.933          & 1.438        & 1.892          & 1.513        & 1.903 \\
FEDformer       & 1.496        & 1.939          & 1.527        & 1.942          & 1.652        & 2.072 \\
\midrule
$\text{TimeMOE}_{small}$    & 1.482        & 2.018          & 1.340        & 1.812          & 1.326        & \second{1.718} \\
$\text{TimeMOE}_{base}$   & 1.482        & 2.018          & \second{1.337} & 1.811          & \second{1.321} & \best{1.715} \\

$\text{Moirai}_{small}$   & 1.627        & 2.095          & 1.671        & 2.062          & 1.665        & 2.018 \\
$\text{Moirai}_{base}$    & 1.627        & 2.095          & 1.605        & 2.023          & 1.574        & 1.947 \\
$\text{Moirai}_{large}$   & 1.629        & 2.097          & 1.637        & 2.049          & 1.594        & 1.970 \\

TimesFM         & 1.597        & 2.125          & 1.415        & 1.898          & 1.387        & 1.807 \\
$\text{Moment}_{small}$   & 1.598        & 2.090          & 1.418        & 1.849          & 1.384        & 1.756 \\
$\text{Moment}_{base}$    & 1.598        & 2.089          & 1.417        & 1.848          & 1.384        & 1.755 \\
$\text{Moment}_{large}$   & 1.597        & 2.088          & 1.416        & 1.846          & 1.382        & 1.754 \\

$\text{Chronos}_{small}$   & 1.519        & 2.184          & 1.397        & 1.915   & 1.380        & 1.845 \\
$\text{Chronos}_{base}$    & 1.517        & 2.184          & 1.394        & 1.902          & 1.376        & 1.841 \\
$\text{Chronos}_{large}$   & 1.508        & 2.161          & 1.384        & 1.896          & 1.377        & 1.832 \\
\midrule
\rowcolor{mediumgray} WindFM (Ours)          & \best{1.434} & 1.991          & \best{1.314} & \best{1.795} & \best{1.274} & 1.744 \\
\bottomrule
\end{tabular}} 
\end{table}

\textbf{3. Cross-Geography Generalization.} To rigorously test the model's robustness and ability to generalize to out-of-distribution data, we conduct an evaluation on a wind power dataset from Inner Mongolia, China. This dataset has a 15-minute resolution and spans from October 1, 2019, to October 30, 2020. We test on forecast horizons of 48, 96, and 144 steps. For the site-specific baseline models, this dataset is partitioned into training, validation, and test sets using a 6:1:3 ratio. In contrast, WindFM and the other time-series foundation models are evaluated directly on the test set in a zero-shot setting, without any training or fine-tuning on the new data. Performance is measured using MAE and RMSE.

\subsection{Zero-Shot Forecasting on Unseen Domestic Sites}

As shown in Tables~\ref{tab:wind_power_full_results} and \ref{tab:zeroshot_full_results}, WindFM achieves lower error than site-specific models like DLinear and TimeXer without requiring target-specific training. It also outperforms larger time-series foundation models such as TimeMOE and Chronos. This suggests that domain-specific pre-training enables WindFM to learn more effective representations for wind power forecasting, an advantage maintained across both seen and unseen time frequencies.

\begin{figure}[t!]
    \centering
    \includegraphics[width=1.0\columnwidth]{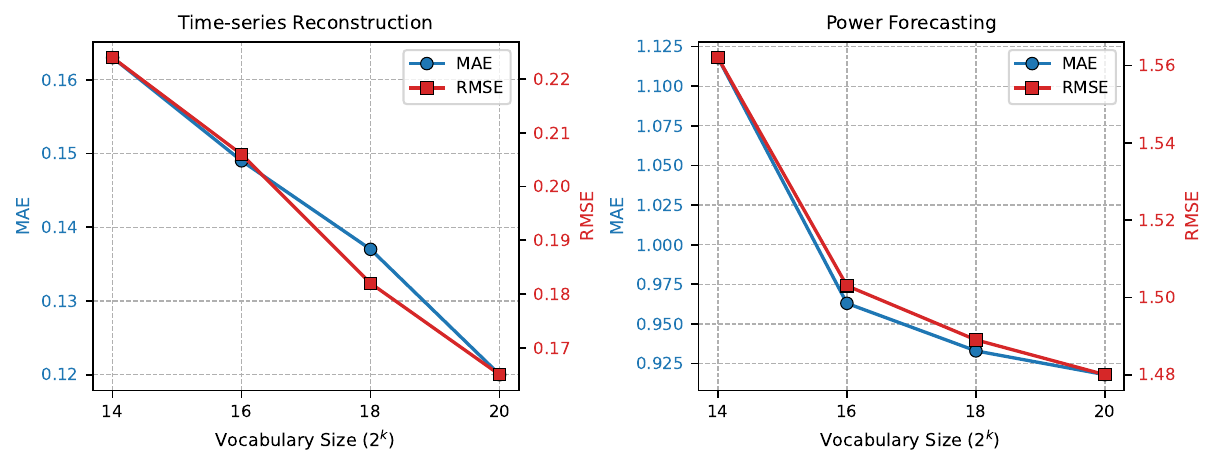}
    \caption{Impact of tokenizer vocabulary size ($2^k$) on performance. Increasing the vocabulary size improves the model's representational capacity, leading to a consistent decrease in both time-series reconstruction error (left) and downstream power forecasting error (right) in terms of MAE and RMSE.
    }
    \label{fig:vocabulary_size_ablation}
\end{figure}

\begin{figure*}[t!]
    \centering
    \includegraphics[width=1.0\textwidth]{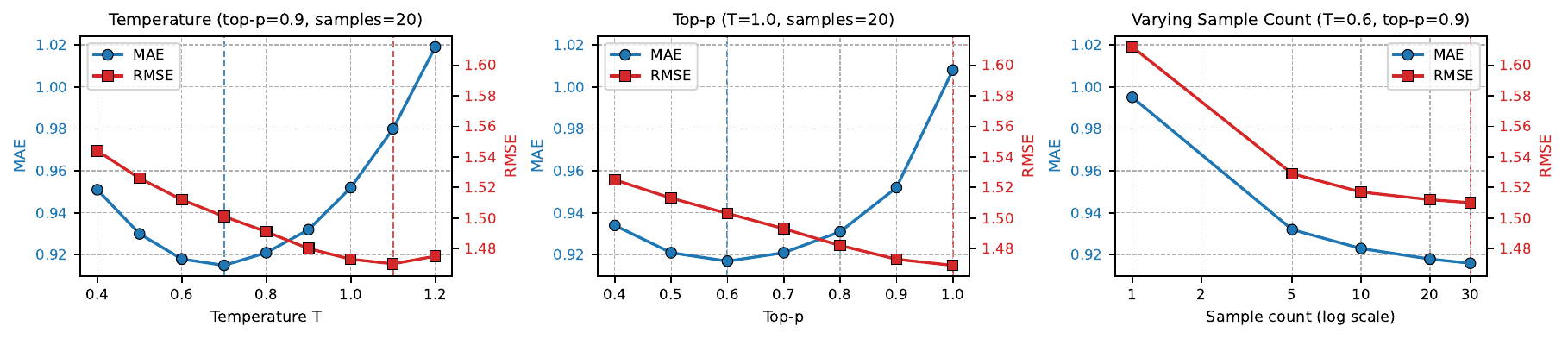}
    \caption{Sensitivity analysis of WindFM's forecasting performance with respect to inference sampling hyperparameters. We evaluate MAE and RMSE by: (a) varying Temperature $T$ (top-$p=0.9$, samples=20); (b) varying top-$p$ ($T=1.0$, samples=20); and (c) varying the sample count ($T=0.6$, top-$p=0.9$). Optimal values are marked by dashed lines.}
    \label{fig:inference_param}
\end{figure*}

\begin{figure*}[t!]
    \centering
    \includegraphics[width=1.0\textwidth]{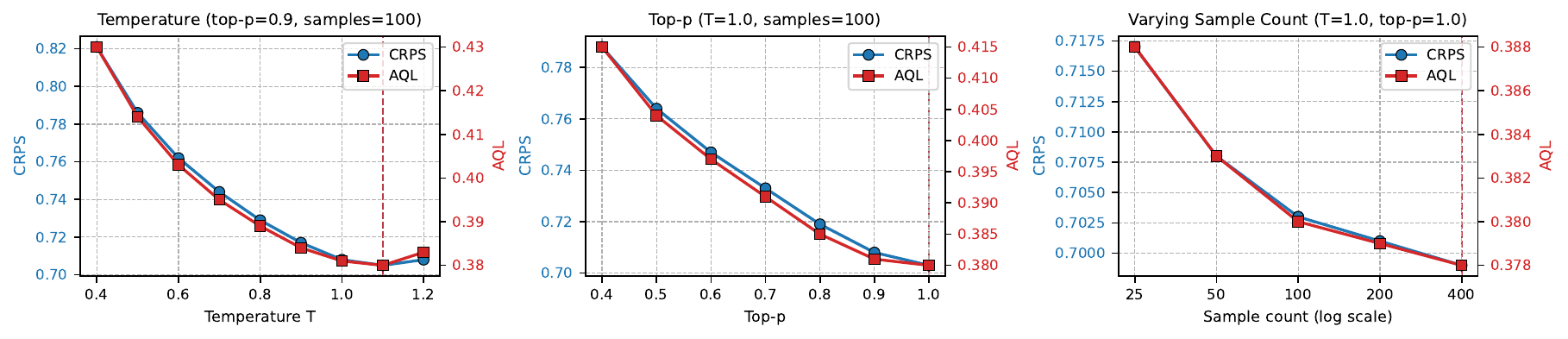}
    \caption{Sensitivity analysis of WindFM's probabilistic forecasting performance with respect to inference sampling hyperparameters. We evaluate CRPS and AQL by: (a) varying Temperature $T$ (top-$p=0.9$, samples=100); (b) varying top-$p$ ($T=1.0$, samples=100); and (c) varying the sample count ($T=1.0$, top-$p=1.0$). Optimal values used for evaluation are marked by dashed lines.}
    \label{fig:inference_param_prob}
\end{figure*}

Figure~\ref{fig:model_param_comparison} further illustrates the trade-off between forecasting performance and model size, reinforcing the key aspects of our approach. The plot positions WindFM in the bottom-left quadrant, visually confirming that it achieves state-of-the-art accuracy (lower MAE and RMSE) while being substantially more parameter-efficient. With only 8.1M parameters, WindFM is an order of magnitude smaller than many time-series foundation models, yet delivers better results. In contrast to the site-specific paradigm, which requires a dedicated model to be trained and tuned for each location, our model delivers high performance in a zero-shot setting without any site-specific adaptation. Taken together, these findings suggest that WindFM's combination of high accuracy and a compact architecture makes it a practical and efficient model for wind power forecasting applications.

\subsection{Cross-Geography Generalization}

To assess the model's robustness and generalization to out-of-distribution data, we evaluate its zero-shot performance on a dataset from Inner Mongolia, China. The results, presented in Table~\ref{tab:my_experiment_results}, show that WindFM maintains a high level of accuracy even when applied to this entirely new geographical region. Notably, WindFM performs comparably to, and in several cases surpasses, site-specific models like TimeXer and PatchTST, which are trained directly on the target dataset. Furthermore, it outperforms other time-series foundation models often by a significant margin. This outcome suggests that the representations learned by WindFM capture transferable physical principles of wind generation, demonstrating the model's generalization capabilities.

\subsection{Probabilistic Forecasting}

The results of the probabilistic forecasting task are in Table~\ref{tab:probabilistic_full_results}. WindFM consistently achieves the lowest CRPS and AQL across all evaluated settings. It outperforms both other foundation models (Chronos, Moirai) and site-specific models adapted for this task with Gaussian outputs. These results highlight the effectiveness of the generative framework for capturing the inherent uncertainty of wind power.

\begin{table}[t!]
\centering
\caption{Ablation study on WindFM's key components.}
\label{tab:ablation_study}
\resizebox{\columnwidth}{!}{%
\begin{tabular}{l c c}
\toprule
\textbf{Configuration Variant} & \textbf{MAE} ($\downarrow$) & \textbf{RMSE} ($\downarrow$) \\
\midrule
\multicolumn{3}{l}{\textit{\textbf{Ablation on Model Architecture \& Prediction Space}}} \\
\rowcolor{gray!15}
\textbf{Discrete Hierarchical (Ours)} & \textbf{0.918} & \textbf{1.480} \\
Discrete Independent & 0.971 & 1.484 \\
Continuous Probabilistic (NLL) & 0.952 & 1.491 \\
Continuous Deterministic (MSE) & 1.237 & 1.744 \\
\midrule
\multicolumn{3}{l}{\textit{\textbf{Ablation on Time Embedding}}} \\
\rowcolor{gray!15}
\textbf{Fourier Embedding (Ours)} & \textbf{0.918} & \textbf{1.480} \\
Time Feature Embedding & 1.078 & 1.560 \\
No Time Information & 1.097 & 1.576 \\
\bottomrule
\end{tabular}%
}
\end{table}

\subsection{Ablation Study}

\begin{figure*}[t!]
    \centering
    \includegraphics[width=1.0\textwidth]{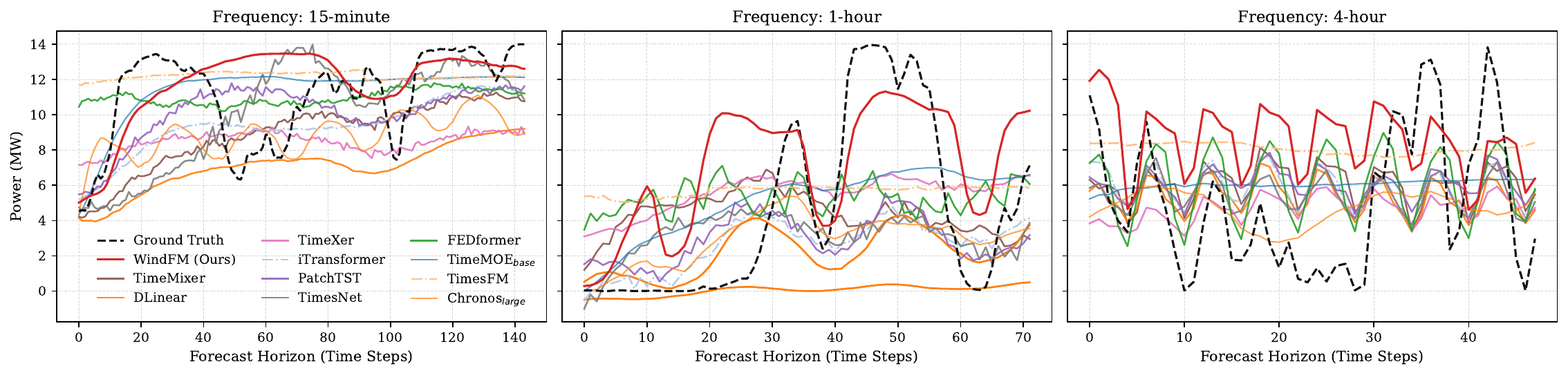}
    \caption{Visualization of WindFM's zero-shot forecasts against other baselines. The plots cover three different time frequencies for an unseen wind farm.}
    \label{fig:prediction_visualize}
\end{figure*}

\begin{figure*}[t!]
    \centering
    \includegraphics[width=1.0\textwidth]{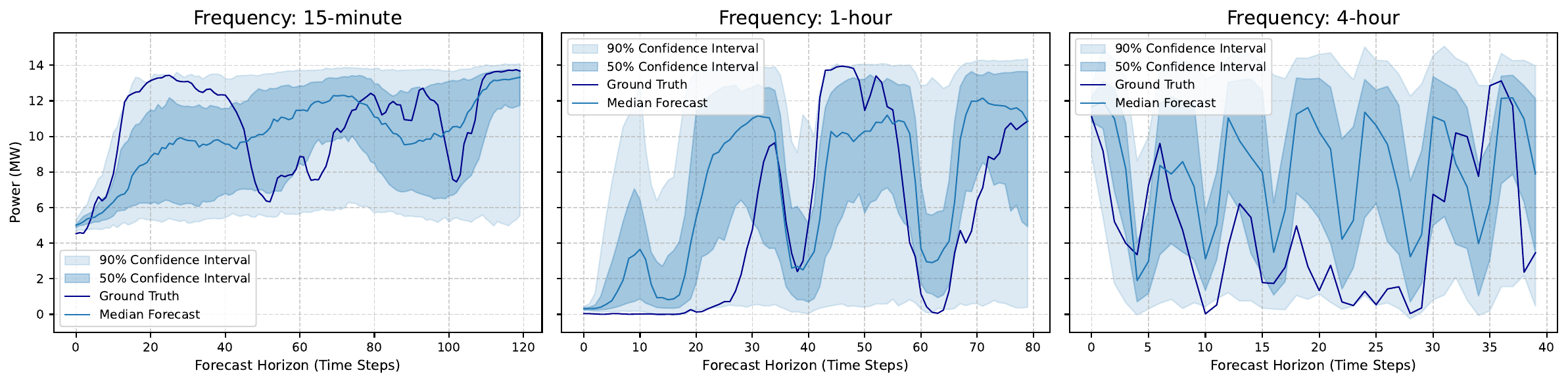}
    \caption{Visualization of WindFM's zero-shot probabilistic forecasts. The plots display the median forecast, alongside the 50\% and 90\% confidence intervals, for an unseen wind farm across three different time frequencies.}
    \label{fig:probabilistic_prediction_visualize}
\end{figure*}

We conduct an ablation study to validate our key design choices (Table~\ref{tab:ablation_study}). First, we examine the architecture and prediction space. The proposed discrete hierarchical framework yields the best performance. A variant predicting coarse and fine subtokens independently (Discrete Independent), ignoring the structured dependency, results in lower accuracy. This confirms the benefit of the coarse-to-fine generation process. Formulating the problem in the continuous domain, either probabilistically (NLL loss) or deterministically (MSE loss), yields inferior results. The substantially larger error of the direct regression (MSE) model affirms the effectiveness of the \emph{discretize-and-generate} paradigm. Second, we assess the temporal encoding. The Fourier-based embedding significantly outperforms a standard learnable time feature embedding. Removing temporal information entirely leads to the worst performance, confirming its critical role.

We also investigate the impact of the tokenizer's vocabulary size, governed by the bit length $k$, on both reconstruction and forecasting performance. As shown in Figure~\ref{fig:vocabulary_size_ablation}, increasing the vocabulary size leads to a monotonic decrease in error for both tasks. A larger vocabulary endows the tokenizer with greater representational capacity, allowing it to capture the nuances of the continuous time series with higher fidelity, which in turn enhances the accuracy of the downstream autoregressive model. However, the gains in forecasting performance exhibit diminishing returns as the vocabulary size increases. Considering this trade-off with the associated growth in model parameters and potential for training instability, we select $k=20$ as a balanced configuration for our final model.

\subsection{Sensitivity to Inference Hyperparameters}

We analyze the sensitivity of forecast quality to key inference hyperparameters: temperature ($T$), top-$p$, and sample count. The results (Figures~\ref{fig:inference_param} and \ref{fig:inference_param_prob}) reveal distinct optimal settings for deterministic and probabilistic tasks.

For deterministic forecasting (Figure~\ref{fig:inference_param}), MAE and RMSE show a U-shaped relationship with temperature and top-$p$, indicating that a balance between determinism and stochasticity is required. Very low values can cause repetitive predictions, while very high values can introduce noise. Performance improves with more samples due to more stable averaging, albeit with diminishing returns.

For probabilistic forecasting (Figure~\ref{fig:inference_param_prob}), optimal settings favor greater sample diversity. CRPS and AQL improve as top-$p$ approaches 1.0. Performance improves as temperature increases to an optimum around $T=1.1$, suggesting that high stochasticity is beneficial for modeling uncertainty, but excessive randomness is detrimental. As before, more samples yield a better empirical distribution. These findings demonstrate the flexibility of the generative framework, allowing inference to be tailored to specific forecasting objectives.

\subsection{Visualization of Forecasting Results}

To provide a qualitative perspective on model performance, we visualize the zero-shot forecasts. Figure~\ref{fig:prediction_visualize} compares the deterministic forecasts from WindFM and baseline models. The plot shows that WindFM's predictions effectively track the complex and volatile patterns of actual power generation, including sharp ramps and high-frequency fluctuations. In contrast, many competing models yield overly smoothed forecasts that fail to capture these critical dynamics.

Figure~\ref{fig:probabilistic_prediction_visualize} showcases WindFM's probabilistic capabilities, displaying the median forecast with 50\% and 90\% confidence intervals. The ground truth consistently lies within the predicted bands, indicating well-calibrated uncertainty estimates. Furthermore, the confidence intervals are adaptive, widening during periods of high volatility and narrowing when the output is more stable, demonstrating the model's ability to dynamically quantify its own predictive uncertainty.

\section{Conclusion}
This paper introduces WindFM, a lightweight and domain-specific foundation model for wind power forecasting. By pre-training a single decoder-only Transformer on hierarchically tokenized data from numerous sites, WindFM learns a universal representation of wind generation dynamics. Our experiments demonstrate state-of-the-art zero-shot performance, outperforming both specialized models trained on target data and larger general-purpose foundation models. The model's strong generalization to new geographical regions, combined with its efficient 8.1M parameter architecture, underscores its suitability for practical deployment. This work validates domain-specific pre-training as an effective alternative to site-by-site modeling, offering a powerful tool for managing renewable-rich power systems.

This work can be extended in three directions. First, we plan to integrate slow-thinking reasoning mechanisms to improve the interpretability of the model's outputs. Second, by leveraging knowledge graphs and retrieval-augmented generation (RAG), the model will be able to dynamically access and exploit historical knowledge from a vast corpus of wind power forecasting data. Finally, we will explore privacy-preserving computation techniques to securely fuse multi-source data, including wind, photovoltaic, and electricity-price datasets, thereby further boosting the model’s generalization capability across the broader energy domain.


%





\ifCLASSOPTIONcaptionsoff
  \newpage
\fi



%

\begin{thebibliography}{10}
\providecommand{\url}[1]{#1}
\csname url@samestyle\endcsname
\providecommand{\newblock}{\relax}
\providecommand{\bibinfo}[2]{#2}
\providecommand{\BIBentrySTDinterwordspacing}{\spaceskip=0pt\relax}
\providecommand{\BIBentryALTinterwordstretchfactor}{4}
\providecommand{\BIBentryALTinterwordspacing}{\spaceskip=\fontdimen2\font plus
\BIBentryALTinterwordstretchfactor\fontdimen3\font minus \fontdimen4\font\relax}
\providecommand{\BIBforeignlanguage}[2]{{%
\expandafter\ifx\csname l@#1\endcsname\relax
\typeout{** WARNING: IEEEtran.bst: No hyphenation pattern has been}%
\typeout{** loaded for the language `#1'. Using the pattern for}%
\typeout{** the default language instead.}%
\else
\language=\csname l@#1\endcsname
\fi
#2}}
\providecommand{\BIBdecl}{\relax}
\BIBdecl

\bibitem{shi2025kronos}
Y.~Shi, Z.~Fu, S.~Chen \emph{et~al.}, ``Kronos: A foundation model for the language of financial markets,'' \emph{arXiv preprint arXiv:2508.02739}, 2025.

\bibitem{zhao2024image}
Y.~Zhao, Y.~Xiong, and P.~Kr{\"a}henb{\"u}hl, ``Image and video tokenization with binary spherical quantization,'' \emph{arXiv preprint arXiv:2406.07548}, 2024.

\bibitem{yu2023language}
L.~Yu, J.~Lezama, N.~B. Gundavarapu \emph{et~al.}, ``Language model beats diffusion--tokenizer is key to visual generation,'' \emph{arXiv preprint arXiv:2310.05737}, 2023.

\bibitem{lu2021review}
P.~Lu, L.~Ye, Y.~Zhao \emph{et~al.}, ``Review of meta-heuristic algorithms for wind power prediction: Methodologies, applications and challenges,'' \emph{Applied Energy}, vol. 301, p. 117446, 2021.

\bibitem{draxl2015wind}
C.~Draxl, A.~Clifton, B.-M. Hodge \emph{et~al.}, ``The wind integration national dataset (wind) toolkit,'' \emph{Applied Energy}, vol. 151, pp. 355--366, 2015.

\bibitem{gwec2024}
\BIBentryALTinterwordspacing
{Global Wind Energy Council}, ``Global wind report 2024,'' GWEC, Tech. Rep., 2024. [Online]. Available: \url{https://gwec.net/global-wind-report-2024/}
\BIBentrySTDinterwordspacing

\bibitem{wwindea2024}
\BIBentryALTinterwordspacing
{World Wind Energy Association}, ``{WWEA} annual report 2024: A challenging year for windpower,'' WWEA, Tech. Rep., 2024. [Online]. Available: \url{https://www.wwindea.org/AnnualReport2024}
\BIBentrySTDinterwordspacing

\bibitem{yang2024survey}
Y.~Yang, H.~Lou, J.~Wu \emph{et~al.}, ``A survey on wind power forecasting with machine learning approaches,'' \emph{Neural Computing and Applications}, vol.~36, no.~21, pp. 12\,753--12\,773, 2024.

\bibitem{huang2023deep}
S.~Huang, C.~Yan, and Y.~Qu, ``Deep learning model-transformer based wind power forecasting approach,'' \emph{Front. Energy Res.}, vol.~10, p. 1055683, 2023.

\bibitem{ansari2024chronos}
A.~F. Ansari, L.~Stella, C.~Turkmen \emph{et~al.}, ``Chronos: Learning the language of time series,'' \emph{arXiv preprint arXiv:2403.07815}, 2024.

\bibitem{shi2024time}
X.~Shi, S.~Wang, Y.~Nie \emph{et~al.}, ``Time-moe: Billion-scale time series foundation models with mixture of experts,'' \emph{arXiv preprint arXiv:2409.16040}, 2024.

\bibitem{zhang2024temporal}
Y.~Hu, H.~Liu, S.~Wu \emph{et~al.}, ``Temporal collaborative attention for wind power forecasting,'' \emph{Applied Energy}, vol. 357, p. 122502, 2024.

\bibitem{liuitransformer}
Y.~Liu, T.~Hu, H.~Zhang \emph{et~al.}, ``itransformer: Inverted transformers are effective for time series forecasting,'' \emph{arXiv preprint arXiv:2310.06625}, 2023.

\bibitem{zhou2022fedformer}
T.~Zhou, Z.~Ma, Q.~Wen \emph{et~al.}, ``Fedformer: Frequency enhanced decomposed transformer for long-term series forecasting,'' in \emph{International Conference on Machine Learning}.\hskip 1em plus 0.5em minus 0.4em\relax PMLR, 2022, pp. 27\,268--27\,286.

\bibitem{wang2024timexer}
Y.~Wang, H.~Wu, J.~Dong \emph{et~al.}, ``Timexer: Empowering transformers for time series forecasting with exogenous variables,'' \emph{Advances in Neural Information Processing Systems}, vol.~37, pp. 469--498, 2024.

\bibitem{nie2022time}
Y.~Nie, N.~H. Nguyen, P.~Sinthong \emph{et~al.}, ``A time series is worth 64 words: Long-term forecasting with transformers,'' \emph{arXiv preprint arXiv:2211.14730}, 2022.

\bibitem{liu2022non}
Y.~Liu, H.~Wu, J.~Wang \emph{et~al.}, ``Non-stationary transformers: Exploring the stationarity in time series forecasting,'' \emph{Advances in neural information processing systems}, vol.~35, pp. 9881--9893, 2022.

\bibitem{wang2024timemixer}
S.~Wang, H.~Wu, X.~Shi \emph{et~al.}, ``Timemixer: Decomposable multiscale mixing for time series forecasting,'' \emph{arXiv preprint arXiv:2405.14616}, 2024.

\bibitem{wu2022timesnet}
H.~Wu, T.~Hu, Y.~Liu \emph{et~al.}, ``Timesnet: Temporal 2d-variation modeling for general time series analysis,'' \emph{arXiv preprint arXiv:2210.02186}, 2022.

\bibitem{zeng2023transformers}
A.~Zeng, M.~Chen, L.~Zhang \emph{et~al.}, ``Are transformers effective for time series forecasting?'' in \emph{Proceedings of the AAAI Conference on Artificial Intelligence}, no.~9, 2023, pp. 11\,121--11\,128.

\bibitem{goswami2024moment}
M.~Goswami, K.~Szafer, A.~Choudhry \emph{et~al.}, ``Moment: A family of open time-series foundation models,'' \emph{arXiv preprint arXiv:2402.03885}, 2024.

\bibitem{das2024decoder}
A.~Das, W.~Kong, R.~Sen \emph{et~al.}, ``A decoder-only foundation model for time-series forecasting,'' in \emph{Forty-first International Conference on Machine Learning}, 2024.

\bibitem{woo2024unified}
G.~Woo, C.~Liu, A.~Kumar \emph{et~al.}, ``Unified training of universal time series forecasting transformers,'' \emph{arXiv preprint arXiv:2402.02592}, 2024.

\bibitem{su2024roformer}
J.~Su, M.~Ahmed, Y.~Lu \emph{et~al.}, ``Roformer: Enhanced transformer with rotary position embedding,'' \emph{Neurocomputing}, vol. 568, p. 127063, 2024.

\bibitem{xiong2020layer}
R.~Xiong, Y.~Yang, D.~He \emph{et~al.}, ``On layer normalization in the transformer architecture,'' in \emph{International conference on machine learning}.\hskip 1em plus 0.5em minus 0.4em\relax PMLR, 2020, pp. 10\,524--10\,533.

\bibitem{zhang2019root}
B.~Zhang and R.~Sennrich, ``Root mean square layer normalization,'' \emph{Adv. Neural Inf. Process. Syst.}, vol.~32, 2019.

\bibitem{loshchilov2017decoupled}
I.~Loshchilov and F.~Hutter, ``Decoupled weight decay regularization,'' \emph{arXiv preprint arXiv:1711.05101}, 2017.

\bibitem{zhou2025empower}
Y.~Zhou and M.~Wang, ``Empower pre-trained large language models for building-level load forecasting,'' \emph{IEEE Transactions on Power Systems}, 2025.

\bibitem{majumder2024exploring}
S.~Majumder, L.~Dong, F.~Doudi \emph{et~al.}, ``Exploring the capabilities and limitations of large language models in the electric energy sector,'' \emph{Joule}, vol.~8, no.~6, pp. 1544--1549, 2024.

\bibitem{wang2024high}
S.~Wang, J.~Shi, W.~Yang \emph{et~al.}, ``High and low frequency wind power prediction based on transformer and bigru-attention,'' \emph{Energy}, vol. 288, p. 129753, 2024.

\bibitem{zhang2021multi}
H.~Zhang, J.~Yan, Y.~Liu \emph{et~al.}, ``Multi-source and temporal attention network for probabilistic wind power prediction,'' \emph{IEEE Transactions on Sustainable Energy}, vol.~12, no.~4, pp. 2205--2218, 2021.

\bibitem{cheng2021augmented}
L.~Cheng, H.~Zang, Y.~Xu \emph{et~al.}, ``Augmented convolutional network for wind power prediction: A new recurrent architecture design with spatial-temporal image inputs,'' \emph{IEEE Transactions on Industrial Informatics}, vol.~17, no.~10, pp. 6981--6993, 2021.

\bibitem{jiang2024eplus}
G.~Jiang, Z.~Ma, L.~Zhang \emph{et~al.}, ``Eplus-llm: A large language model-based computing platform for automated building energy modeling,'' \emph{Appl. Energy}, vol. 367, p. 123431, 2024.

\bibitem{wen2022continuous}
H.~Wen, P.~Pinson, J.~Ma \emph{et~al.}, ``Continuous and distribution-free probabilistic wind power forecasting: A conditional normalizing flow approach,'' \emph{IEEE Transactions on Sustainable Energy}, vol.~13, no.~4, pp. 2250--2263, 2022.

\bibitem{li2025novel}
S.~Li, X.~Li, Y.~Jiang \emph{et~al.}, ``A novel frequency-domain physics-informed neural network for accurate prediction of 3d spatio-temporal wind fields in wind turbine applications,'' \emph{Applied Energy}, vol. 386, p. 125526, 2025.

\bibitem{lai2024bert4st}
Z.~Lai, T.~Wu, X.~Fei \emph{et~al.}, ``Bert4st:: Fine-tuning pre-trained large language model for wind power forecasting,'' \emph{Energy Conv. Manag.}, vol. 307, p. 118331, 2024.

\bibitem{yu2025pricefm}
R.~Yu, C.~Gu, J.~Stiasny \emph{et~al.}, ``Pricefm: Foundation model for probabilistic electricity price forecasting,'' \emph{arXiv preprint arXiv:2508.04875}, 2025.

\bibitem{shahid2021novel}
F.~Shahid, A.~Zameer, and M.~Muneeb, ``A novel genetic lstm model for wind power forecast,'' \emph{Energy}, vol. 223, p. 120069, 2021.

\bibitem{abedinia2024wind}
O.~Abedinia, A.~Ghasemi-Marzbali, M.~Shafiei \emph{et~al.}, ``Wind power forecasting enhancement utilizing adaptive quantile function and cnn-lstm: a probabilistic approach,'' \emph{IEEE Transactions on Industry Applications}, vol.~60, no.~3, pp. 4446--4457, 2024.

\bibitem{zhu2025novel}
J.~Zhu and Y.~He, ``A novel photovoltaic power probabilistic forecasting model based on monotonic quantile convolutional neural network and multi-objective optimization,'' \emph{Energy Conversion and Management}, vol. 323, p. 119219, 2025.

\end{thebibliography}


\end{document}